%% file: main.tex
\documentclass[sigconf]{acmart}
\usepackage{subcaption}
\usepackage{booktabs}
\usepackage{multirow}
\usepackage{multicol}
\usepackage{xcolor}
\usepackage{longtable}
\usepackage[document]{ragged2e}

\AtBeginDocument{%
    \justifying
  }

\copyrightyear{2026}
\acmYear{2026}
\setcopyright{cc}
\setcctype{by}
\acmConference[KDD 2026] {Proceedings of the 32nd ACM SIGKDD Conference on Knowledge Discovery and Data Mining V.2}{August 9--13, 2026}{Jeju Island, Republic of Korea.}
\acmBooktitle{Proceedings of the 32nd ACM SIGKDD Conference on Knowledge Discovery and Data Mining V.2 (KDD 2026), August 9--13, 2026, Jeju Island, Republic of Korea}
\acmISBN{979-8-4007-2259-2/2026/08}
\acmDOI{10.1145/3770855.3817566}
\begin{document}

\title{CarbonFluxBench: A Global Benchmark for Upscaling of Carbon Fluxes Using Zero-Shot Learning}

\author{Aleksei Rozanov}
\email{rozan012@umn.edu}
\orcid{0009-0004-6285-7066}
\affiliation{%
  \institution{Department of Computer Science \& Engineering, University of Minnesota -- Twin Cities}
  \city{Minneapolis}
  \state{MN}
  \country{USA}
}

\author{Arvind Renganathan}
\email{renga@umn.edu}
\affiliation{%
  \institution{Department of Computer Science \& Engineering, University of Minnesota -- Twin Cities}
  \city{Minneapolis}
  \state{MN}
  \country{USA}
}

\author{Yimeng Zhang}
\email{zhan8460@umn.edu}
\affiliation{%
  \institution{Department of Computer Science \& Engineering, University of Minnesota -- Twin Cities}
  \city{Minneapolis}
  \state{MN}
  \country{USA}
}

\author{Vipin Kumar}
\email{kumar001@umn.edu}
\affiliation{%
  \institution{Department of Computer Science \& Engineering, University of Minnesota -- Twin Cities}
  \city{Minneapolis}
  \state{MN}
  \country{USA}
}


\input{abstract}


\vspace{-0.2cm}

\maketitle

\input{intro_v4}

\input{problem_formulation}
\input{related}
\input{method}

\input{eval}

\input{results}
\input{discussion}

\input{conclusion}

\begin{acks}
This work was supported by the NSF LEAP Science and Technology Center (Award 2019625) and the NSF grant (2313174).
Computational resources were provided by the Minnesota Supercomputing Institute (MSI).  A.
Renganathan acknowledges support from the University of Minnesota Doctoral Dissertation Fellowship (2025–26). We would like to acknowledge the site teams produced EC data used in this study distributed by FLUXNET, ICOS, AmeriFlux, and JapanFlux. AmeriFlux is supported by the U.S. Department of Energy Office of Science.
ERA5-Land reanalysis data were produced by the ECMWF Copernicus Climate Change Service. MODIS data were obtained from the NASA Land Processes Distributed Active Archive Center (LP DAAC).
\end{acks}

\bibliographystyle{ACM-Reference-Format}
\bibliography{ref}

\appendix

\input{appendix_a}
\end{document}

%% file: abstract.tex
\vspace{-1cm}
\begin{abstract}
Accurately quantifying terrestrial carbon exchange is essential for climate policy and carbon accounting, yet models must generalize to ecosystems underrepresented in sparse eddy covariance observations. Despite this challenge being a natural instance of zero-shot spatial transfer learning for time series regression, no standardized benchmark exists to rigorously evaluate model performance across geographically distinct climate regimes and vegetation types.

We introduce \textit{CarbonFluxBench}, the first benchmark for zero-shot spatial transfer in carbon flux upscaling. \textit{CarbonFluxBench} comprises over 1.1 million daily observations from 474 flux tower sites globally (2000–2024). It provides: (1) stratified evaluation protocols that explicitly test generalization across unseen vegetation types and climate regimes, separating spatial transfer from temporal autocorrelation; (2) a harmonized set of remote sensing and meteorological features to enable flexible architecture design; and (3) baselines ranging from tree-based methods to domain-generalization architectures. By bridging machine learning methodologies and Earth system science, \textit{CarbonFluxBench} aims to enable systematic comparison of transfer learning methods, serves as a testbed for regression under distribution shift, and contributes to the next-generation climate modeling efforts.
\end{abstract}

%% file: intro_v4.tex
\section{Introduction}
As highlighted by the 2023 Intergovernmental Panel on Climate Change (IPCC) assessment \cite{masson2023summary}, anthropogenic $\mathrm{CO_2}$ emissions have unequivocally driven global temperature increases across atmosphere, land, and ocean. While numerous carbon removal approaches are under development \cite{smith2024state, fawzy2021industrial, babakhani2022potential}, biological sequestration, i.e. the photosynthetic capture and storage of $\mathrm{CO_2}$ as biomass and soil organic carbon, remains one of the few scalable natural mechanisms for  atmospheric decarbonization.

Eddy covariance (EC) {\cite{baldocchi2003assessing}} towers provide an on-ground instrumented way to quantify this process through measuring net ecosystem exchange (NEE), representing land–atmosphere carbon flux within a footprint extending up to roughly 3,000 m around the tower {\cite{chu2021representativeness}}. Despite their widespread use, EC observations remain geographically sparse due to high operational costs, technical complexity, and inaccessibility of certain regions (e.g., tropical rainforests).  Furthermore, EC data is publicly available only from several hundred sites, with a combined spatial footprint of flux towers at less than 0.015\% of the Earth's terrestrial surface. Thus, despite high-fidelity, these observations are insufficient to produce spatially continuous assessments of global carbon sequestration needed for climate policy, carbon accounting, and calibration and validation of Earth system models (e.g., CESM, CMIP).

Researchers have therefore developed regional and global-scale datasets by posing the \textbf{upscaling} problem \cite{ge2019principles}, i.e., a \textit{bottom-up} task that infers continuous flux fields from point-level EC observations using gridded covariate data. Several global products including FLUXCOM \cite{jung2020scaling, nelson2024x}, GloFlux\cite{yuan2025global} and others \cite{zeng2020global}  have been produced using classical ML approaches based on regression trees and ensembling, with growing interest in deep learning and transfer learning approaches \cite{nathaniel2023metaflux}.

\begin{figure*}[t]
    \centering
    \includegraphics[width=0.45\textwidth]{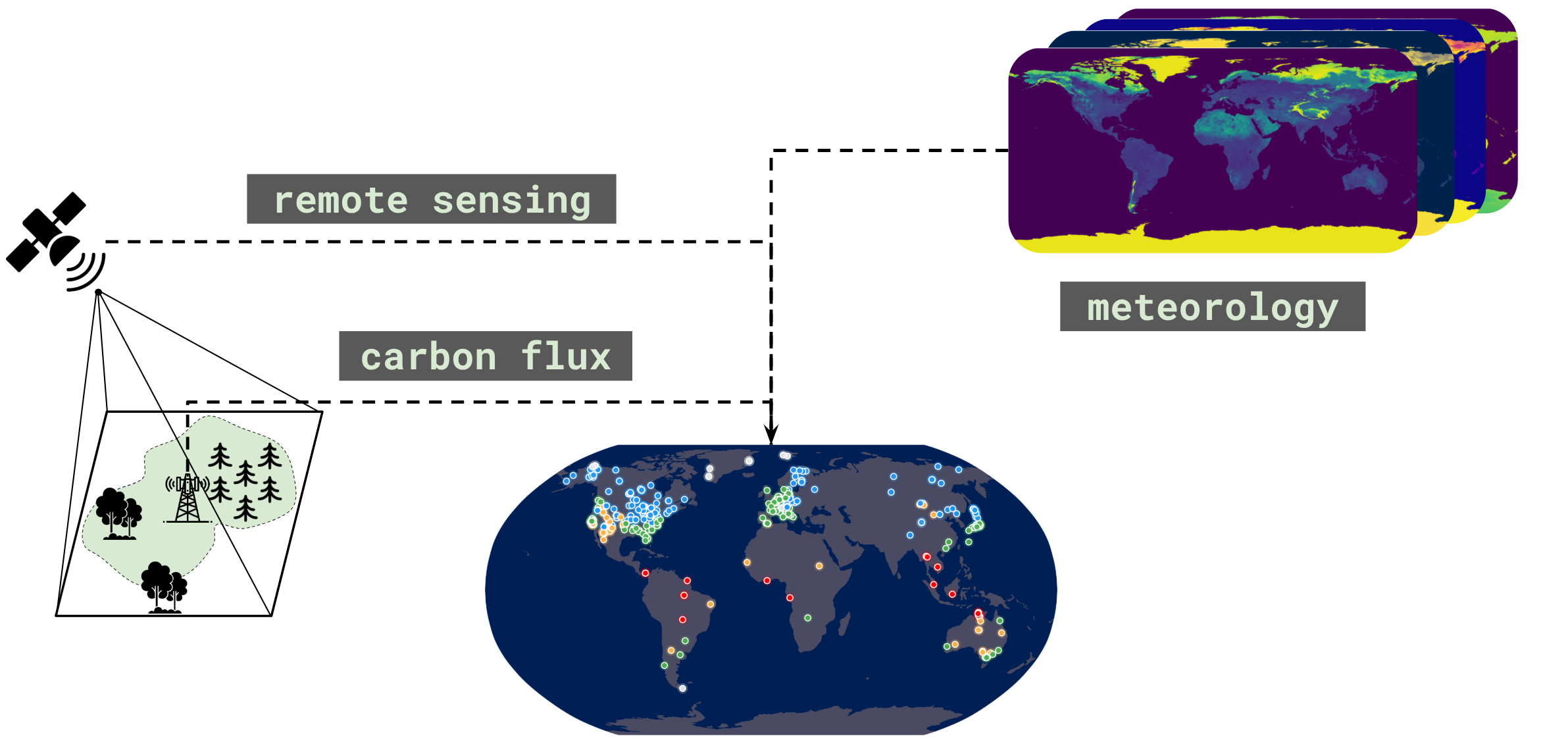}
    \vspace{-0.2cm}
    \caption{\small Overview of the \textit{CarbonFluxBench} setup. Eddy covariance flux observations provide sparse footprint-level targets, while remote sensing and meteorological data serve as gridded inputs for spatial generalization.}
    \vspace{-0.3cm}
    \label{fig:CarbonFluxBench-overview}
\end{figure*}

From a machine learning perspective, this setting naturally corresponds to a \textbf{zero-shot spatial transfer learning problem}, where models must generalize to geographically distinct locations with no prior observations. Each geographical location represents a distinct domain with its own climate regime, vegetation type, and environmental characteristics, making this a natural instance of domain generalization. The model must learn mappings from meteorological and remote sensing (RS) features, along with site-specific metadata such as vegetation type and climate class, to predict carbon fluxes at entirely new locations without any labeled flux data. 

However, carbon flux upscaling remains particularly challenging due to strong spatial heterogeneity, non-uniform data availability and quality.  Flux–feature relationships are highly context-dependent; e.g.,  the same satellite-derived vegetation index and meteorological conditions may lead to different carbon flux values at different locations.  As a result, models trained on well-observed locations often perform well in-distribution but degrade sharply when applied to new geographical regions. Compounding to this challenge, observational coverage is most sparse in regions critical to the global carbon cycle: 1)  tropical ecosystems (high productivity, year-round growing seasons disproportionately contributing to the carbon exchange) and 2) high-latitude regions (that are undergoing rapid warming and permafrost thaw leading to critical shifts in the global carbon cycle). Robust spatial transfer learning is therefore essential for global-scale flux estimation.

There is a growing interest in both ML and domain science communities in taking advantage of advances in machine learning, many of which have been developed in related contexts across entirely different problem domains. As an example, recent works have increasingly emphasized domain generalization and task-aware learning strategies that aim to separate location-specific variability from shared ecosystem dynamics \cite{renganathan2025task, nathaniel2023metaflux}. However, progress in carbon flux upscaling remains constrained by methodological gaps on the machine learning side:

\textbf{Gap I.} Transfer learning for time series regression remains underrepresented in the ML research landscape. Unlike classification, \textbf{regression under distribution shift} is dominated by systematic errors rather than discrete misclassifications: shifts in scale, variance, or noise structure across domains can lead to biased predictions even when inputs appear similar. Continuous domain heterogeneity further alters input–target relationships, making distribution shift difficult to detect and correct. Addressing these challenges requires: (1) absolute and scale-normalized metrics to detect magnitude-dependent biases, (2) stratified evaluation to prevent performance in dominant domains from masking failures in underrepresented regions, and (3) baselines suited to  continuous targets rather than classification. No existing standardized evaluation framework for time series or classification-oriented transfer learning provides this combination.

\textbf{Gap II.} Most existing frameworks for time series regression and forecast primarily assess temporal generalization (i.e., temporal extrapolation) at fixed locations (e.g., Monash Archive \cite{godahewa2021monash}, Informer benchmarks \cite{zhou2021informer}). They do not address spatial generalization across geographically distinct locations with different environmental contexts, despite this being the natural problem structure for many scientific applications. This limitation is particularly critical for carbon flux upscaling, where the most important biomes (tropical, high-latitude) are precisely those with weakest observational coverage. Addressing this gap requires:  (a) stratified evaluation across climate zones and vegetation types, and (b) standardized protocols for comparing domain generalization and transfer learning architectures. 

\textbf{Gap III.} Carbon flux upscaling has remained largely isolated from mainstream machine learning research, limiting the adoption of recent advances in representation learning, transfer learning, and generalization. At the same time, spatial transfer learning for time series regression remains largely underexplored within the ML community, despite rapid progress in domain generalization and adaptation methods. This disconnect has slowed cross-pollination between the two fields, leaving a clear opportunity to align scientifically meaningful problems with active ML research directions.

However, these gaps (domain and ML) make meaningful comparison across methods for carbon upscaling difficult, as existing studies differ substantially in their choice of flux tower sites, input features, target variables, and evaluation approaches, making reported improvements difficult to reproduce, compare, and interpret. To address these challenges, we introduce \textit{CarbonFluxBench}, the first comprehensive transfer learning benchmark for upscaling terrestrial $\mathrm{CO_2}$ fluxes, simultaneously advancing scientific modeling of carbon flux cycles and the methodological development of spatial transfer learning for time series regression (a natural instance of domain generalization and adaptation). \textit{CarbonFluxBench}\footnote{Code: https://github.com/alexxxroz/CarbonFluxBench} enables systematic evaluation of diverse model architectures under zero-shot spatial generalization settings and comprises the following components:

\begin{itemize}
\item A harmonized set of meteorological and remote sensing features at daily temporal resolution for each flux tower site (enables standardized comparison across studies);
\item A dedicated Python library for building highly customizable workflows, supporting data analysis, feature engineering and selection, efficient data loading, model training using \textit{PyTorch} \cite{paszke2019pytorch}, and standardized evaluation;
\item Benchmarked state-of-the-art machine learning models evaluated using regression-appropriate metrics (R², RMSE, normalized MAE) under a rigorous zero-shot evaluation protocol, establishing baseline performance for future comparisons;
\item Two comprehensive train--test splits stratified by both climate and vegetation type to expose scale-dependent biases across domains and comprehensively assess global model generalization.
\end{itemize}

More importantly, \textit{CarbonFluxBench} enables investigation of previously underexplored questions about spatial transfer learning for carbon flux upscaling: 
\begin{enumerate}
    \item How do different neural architectures handle distribution shifts in time series regression?
    \item Do models explicitly designed for transfer learning exhibit better spatial generalization than standard supervised approaches?
    \item Can domain generalization and adaptation architectures be successfully used in the zero-shot spatial transfer learning setting?
    \item Which meta-learning strategies are most effective for spatial  adaptation? 
    \item Can models learn ecosystem dynamics that generalize across geographical regions, and does this improve performance in poorly observed biomes?
\end{enumerate}

%% file: problem_formulation.tex
\section{Problem Formulation}
\subsection{Carbon Flux Upscaling as Spatial Regression}

Let $\Omega \subset \mathbb{R}^2$ denote the spatial domain (e.g., the Earth's land surface) and let $\mathcal{T} = \{1, 2, \ldots, T\}$ represent the temporal domain with daily resolution. We are given a set of eddy covariance tower observations:
\begin{equation}
\mathcal{D} = \{(s_i, \mathbf{z}_i, \mathbf{x}_{i,t}, \mathbf{y}_{i,t}, t) \mid i = 1, \ldots, N; \, t \in \mathcal{T}_i\}
\end{equation}
\vspace{-0.2cm}
where:
\begin{itemize}
    \item $s_i = (\text{lat}_i, \text{lon}_i) \in \Omega$ is the spatial coordinates of site $i$;
    \item $\mathbf{z}_i$ represents site-specific metadata;
    \item $\mathbf{x}_{i,t} \in \mathbb{R}^F$ is the feature vector (drivers) at site $i$ and time $t$, comprising:
    \begin{equation}
    \mathbf{x}_{i,t} = [\mathbf{x}^{\text{RS}}_{i,t}, \, \mathbf{x}^{\text{meteo}}_{i,t}]
    \end{equation}
    where $\mathbf{x}^{\text{RS}}_{i,t}$ are remote sensing features, $\mathbf{x}^{\text{meteo}}_{i,t}$ are meteorological drivers;
    \item $\mathbf{y}_{i,t} \in \mathbb{R}^M$ is the target carbon flux observation at site $i$ and time $t$, where $M$ is the number of different fluxes to model;
    \item $\mathcal{T}_i \subseteq \mathcal{T}$ is the set of observation days available for site $i$.
\end{itemize}

\textbf{The upscaling task} accounts for spatial heterogeneity by conditioning predictions on both time-varying features and site-specific ecosystem context. The model learns a function $f_{\boldsymbol{\theta}}: \mathbb{R}^F \times \mathcal{Z} \rightarrow \mathbb{R}^M$ that maps features and site metadata to carbon fluxes:
\begin{equation}
\hat{\mathbf{y}}_{i,t} = f_{\boldsymbol{\theta}}(\mathbf{x}_{i,t}, \mathbf{z}_i)
\end{equation}
This formulation explicitly recognizes that the relationship between drivers ($\mathbf{x}_{i,t}$) and fluxes ($\mathbf{y}_{i,t}$) varies systematically across ecosystems and climates. 

Some models can explicitly leverage temporal dependencies by operating on a sliding window of past observations. For such models, the input at time $t$ consists of a sequence of $W$ consecutive days:
\begin{equation}
\mathbf{X}_{i,t}^{(W)} = [\mathbf{x}_{i,t-W+1}, \mathbf{x}_{i,t-W+2}, \ldots, \mathbf{x}_{i,t}] \in \mathbb{R}^{W \times F}
\end{equation}
and the prediction is made based on this temporal context:
\begin{equation}
\hat{\mathbf{y}}_{i,t} = f_{\boldsymbol{\theta}}(\mathbf{X}_{i,t}^{(W)}, \mathbf{z}_i)
\end{equation}

In \textit{CarbonFluxBench}, we use $W = 30$ days with a stride of 15 days for training of all the models with temporal context windows.

\subsubsection{Zero-Shot Spatial Transfer}

CarbonFluxBench explicitly evaluates spatial generalization through a hold-out strategy that partitions sites rather than time steps. Let $\mathcal{S} = \{s_1, s_2, \ldots, s_N\}$ denote the set of all flux tower sites. We define a partition:
\begin{equation}
\mathcal{S} = \mathcal{S}_{\text{train}} \cup \mathcal{S}_{\text{test}}, \quad \mathcal{S}_{\text{train}} \cap \mathcal{S}_{\text{test}} = \emptyset
\end{equation}

\textbf{Zero-shot transfer} evaluates a model's ability to generalize to completely unseen geographical locations:
\begin{equation}
\text{Train: } \mathcal{D}_{\text{train}} = \{(\mathbf{x}_i, \mathbf{z}_i, \mathbf{y}_{i,t}, t) \mid i \in \mathcal{S}_{\text{train}}, \, t \in \mathcal{T}_i\}
\end{equation}
\begin{equation}
\text{Test: } \mathcal{D}_{\text{test}} = \{(\mathbf{x}_j, \mathbf{z}_j, \mathbf{y}_{j,t}, t) \mid j \in \mathcal{S}_{\text{test}}, \, t \in \mathcal{T}_j\}
\end{equation}

During training, the model has access to feature vectors $\boldsymbol{\phi}(\mathbf{x}, t)$ for all spatiotemporal locations, but carbon flux labels $\mathbf{y}_{j,t}$ are withheld for test sites $j \in \mathcal{S}_{\text{test}}$. At inference, the model must predict fluxes at test locations using only the learned representations from training sites and the available features at test sites.





\subsubsection{Stratified Evaluation}

To ensure comprehensive assessment across diverse ecosystems and climates, \textit{CarbonFluxBench} provides two stratified train-test splits:

\textbf{IGBP-stratified split}: Sites are partitioned based on the International Geosphere–Biosphere Programme (IGBP) vegetation type $v \in \mathcal{V}$ (16 classes: croplands, forests, grasslands, etc.), ensuring representation of all major ecosystem types:
%
%
with approximately 80\% of sites from each vegetation type allocated to training.

\textbf{Köppen-stratified split}: Sites are partitioned based on Köppen-Geiger climate class \cite{essd-13-5087-2021}, $c \in \mathcal{C}$ (5 main classes: tropical, arid, temperate, continental, polar).
%

This stratification enables fine-grained analysis of model performance across different environmental conditions and vegetation types, revealing which ecosystems or climates pose the greatest transfer learning challenges.







%% file: related.tex
\section{Related Work}

\textbf{Carbon flux upscaling efforts.} The primary goal of carbon flux upscaling is to produce gridded estimates of terrestrial carbon exchange from sparse eddy covariance (EC) observations for climate modeling and decision-making. Most approaches model gross primary production (GPP), ecosystem respiration (RECO), and net ecosystem exchange (NEE) as functions of remotely sensed vegetation indicators and meteorological drivers. Notable operational products include FLUXCOM \cite{jung2020scaling} and FLUXCOM-X-BASE \cite{nelson2024x}, which combine multiple algorithmic families and multi-modal data to generate global flux estimates. Recent efforts have explored meta-learning \cite{nathaniel2023metaflux}, transfer learning with plant functional type specialization \cite{yuan2025global}, and knowledge-guided approaches incorporating physics constraints \cite{fan2025estimating}. Despite these advances, which can serve as stepping stones for new research, existing studies differ substantially in their selection of sites, features, and evaluation protocols, making systematic comparison difficult. 

\textbf{Transfer learning benchmarks in machine learning.} While transfer learning has become central to modern ML, existing benchmarks exhibit critical gaps. Domain adaptation benchmarks (Office-31 \cite{saenko2010adapting}, VisDA \cite{peng2017visda}, DomainNet \cite{peng2019moment}) and NLP evaluation suites (GLUE \cite{wang2018glue}) predominantly focus on classification tasks. Transfer learning for regression, particularly time series regression remains underexplored. Existing time series benchmarks \cite{godahewa2021monash, zhou2021informer, wu2021autoformer} evaluate temporal generalization (forecasting future values at known locations) rather than spatial transfer across geographically distinct domains, 

No benchmark systematically evaluates spatial generalization for time series regression, which happens to be of great importance in Earth system sciences \cite{reichstein2019deep, camps2021deep}, ecology \cite{karpatne2017machine}, and other scientific domains where observations are spatially sparse yet global predictions are needed. Scientific applications offer natural testbeds for spatial transfer learning, featuring genuine heterogeneity, interpretable physical features, and high societal impact, and yet this opportunity remains largely unexploited in the machine learning benchmarking ecosystem \cite{willard2022integrating, karpatne2017machine}.

\textbf{CarbonFluxBench contribution.} CarbonFluxBench directly addresses these gaps by providing the first benchmark for zero-shot spatial transfer learning in time series regression, grounded in the scientifically and societally critical problem of terrestrial carbon flux upscaling. By standardizing site selection, feature sets, and evaluation protocols, CarbonFluxBench enables systematic comparison of transfer learning methods while bridging machine learning benchmarks with real-world scientific discovery. A comprehensive discussion of existing transfer learning benchmarks and their limitations is provided in the expanded version in arXiv.

%% file: method.tex
\section{Methodology}
\subsection{Eddy Covariance}

The ground truth in \textit{CarbonFluxBench} is derived from eddy covariance measurements obtained through major global and regional networks, including FLUXNET \cite{baldocchi2001fluxnet}, AmeriFlux \cite{novick2018ameriflux}\footnote{The AmeriFlux data policy requires individual site citations. Per KDD 2026 proceedings policy limiting references and appendices to 3 pages, the complete list of AmeriFlux site citations is provided in the expanded version of this paper at arXiv.}, ICOS \cite{rebmann2018icos}, and JapanFlux \cite{ueyama2025japanflux2024}. The benchmark aggregates daily time series from 474 flux tower sites covering the period 2000–2024. These sites span all five main Köppen–Geiger climate classes (Fig.~\ref{fig:sites}) and nearly all ecoregions defined by IGBP classification, excluding only the barren category.

Because EC data are inherently noisy, intermittent, and subject to site-specific post processing, we rely on fluxes standardized under the ONEFlux methodology \cite{pastorello2020fluxnet2015}. This pipeline partitions the observed net flux into three key components, which serve as prediction targets in \textit{CarbonFluxBench}:
\begin{itemize}
    \item GPP (GPP\_NT\_VUT\_USTAR50): gross primary production, representing downward carbon flux due to photosynthesis; positive during daytime and zero at night. 
    \item RECO (RECO\_NT\_VUT\_USTAR50): ecosystem respiration, representing upward flux; positive at night and zero during the day. 
    \item NEE (NEE\_VUT\_USTAR50): net ecosystem exchange, the balance between GPP and RECO; negative during daytime (uptake) and positive at night (release). 
\end{itemize}

We also include NEE\_VUT\_USTAR50\_QC (Fig.~\ref{fig:qc}), a continuous binary quality-control flag (0 = lowest, 1 = highest quality). Higher QC values indicate a larger fraction of valid hourly observations contributing to the daily aggregate.



\subsection{Remote Sensing Data}

Vegetation properties are among the strongest determinants of terrestrial carbon fluxes, and satellite-based RS enables systematic monitoring of these characteristics at the global scale. For \textit{CarbonFluxBench}, we use data from NASA’s Moderate Resolution Imaging Spectroradiometer (MODIS) onboard the Terra satellite. 

The product used here is MOD09GA \cite{mod09ga}, which provides estimates of surface spectral reflectance corrected for atmospheric scattering and absorption at 500 m spatial and daily temporal resolution. We extracted seven spectral bands (Red, NIR, Blue, Green, SWIR1–3) and the corresponding cloud mask using Google Earth Engine (GEE) \cite{gorelick2017google}. For each flux tower site, a 2 km × 2 km window was centered on the tower location, and mean reflectance values were computed. The cloud mask was converted into a binary variable (1 = cloud, 0 = clear), then averaged over the window to yield a continuous cloud fraction.  Due to the lack of band-level sensor error estimates in MOD09GA, cloud fraction is the only available proxy for RS data quality in the given benchmark.

\subsection{Meteorological Drivers}
Meteorological conditions strongly influence terrestrial carbon dynamics by regulating photosynthesis, respiration, and evapotranspiration. CarbonFluxBench incorporates climate variables from the ERA5-Land reanalysis \cite{munoz2021era5}. ERA5-Land provides hourly estimates at 0.1° spatial resolution globally.

For each flux tower site, a 2 km x 2 km buffer is constructed and centered on the site coordinates. The overlapping ERA5-Land grid cells within this buffer are averaged and used to derive meteorological predictors. A small subset of coastal sites (10 in total) lacks valid ERA5-Land pixels within the initial buffer. For these sites, the buffer is adaptively expanded to identify the nearest available grid cell, leveraging the spatial smoothness inherent to reanalysis products and minimizing information loss.

From the selected grid cells, we derive all available ERA5-Land variables provided through GEE, covering radiation, moisture, temperature, wind, and surface energy balance processes, resulting in 150 meteorological drivers. 

\subsection{Data Harmonization}
The final number of the EC sites (474) in \textit{CarbonFluxBench} includes only sites with complete availability of all predictor variables and target fluxes. Sites and observations prior to February 2000 were excluded, as MODIS became operational only after that date. All predictors were harmonized to daily resolution: ERA5-Land data summed or averaged per day, MOD09GA gap-filled for every site. Each sample combines MODIS reflectance, ERA5-Land meteorology, IGBP type reported by the site team and Köppen climate class with corresponding carbon flux targets. Continuous predictors and targets were standardized using z-score normalization (computed on the training split only).

\begin{figure}
    \centering
    \includegraphics[width=1\linewidth]{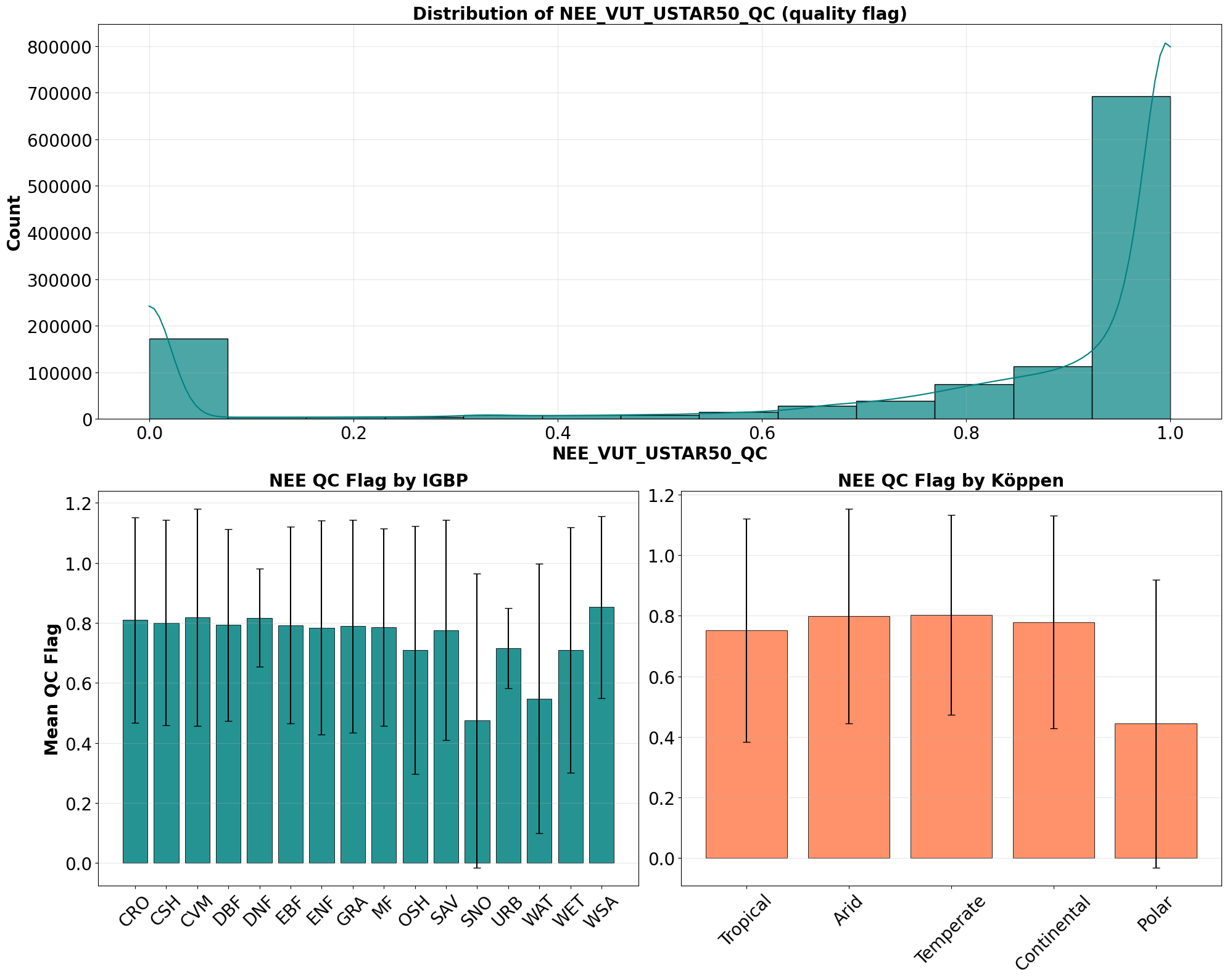}
    \caption{\small Summary of data sample quality in \textit{CarbonFluxBench}. The \textit{upper panel} shows the empirical distribution of $NEE\_VUT\_USTAR50\_QC$ across all samples. The \textit{lower left panel} reports mean quality values aggregated by IGBP class, with bars indicating one standard deviation. The \textit{lower right panel} shows the same aggregation for K\"{o}ppen climate classes.}
    \vspace{-1.cm}
    \label{fig:qc}
\end{figure}

\subsection{Data Statistics}
In total, \textit{CarbonFluxBench} comprises 3 target variables and one resembling quality-control flag, 150 ERA5-Land features, 12 MODIS features, 2 K\"{o}ppen climate variables (long and short), and 1 IGBP vegetation type, in addition to site latitude, longitude, and time of observation. The benchmark contains \textbf{1,170,682} timestamps, of which \textbf{490,232} correspond to the highest-quality observations ($NEE\_VUT\_USTAR50\_QC = 1$).

Figure~\ref{fig:qc} illustrates the empirical distribution of samples across quality-control values, revealing two pronounced modes near $0$ and $1$. We further report the mean quality flag and its standard deviation stratified by IGBP and K\"{o}ppen classes. Overall, quality is relatively uniform across categories; however, SNO (permanent snow and ice) and WAT (water bodies) exhibit the lowest mean quality among IGBP classes. Among K\"{o}ppen climates, Polar regions show the lowest sample quality.

\begin{figure}
    \centering
    \includegraphics[width=0.9\linewidth]{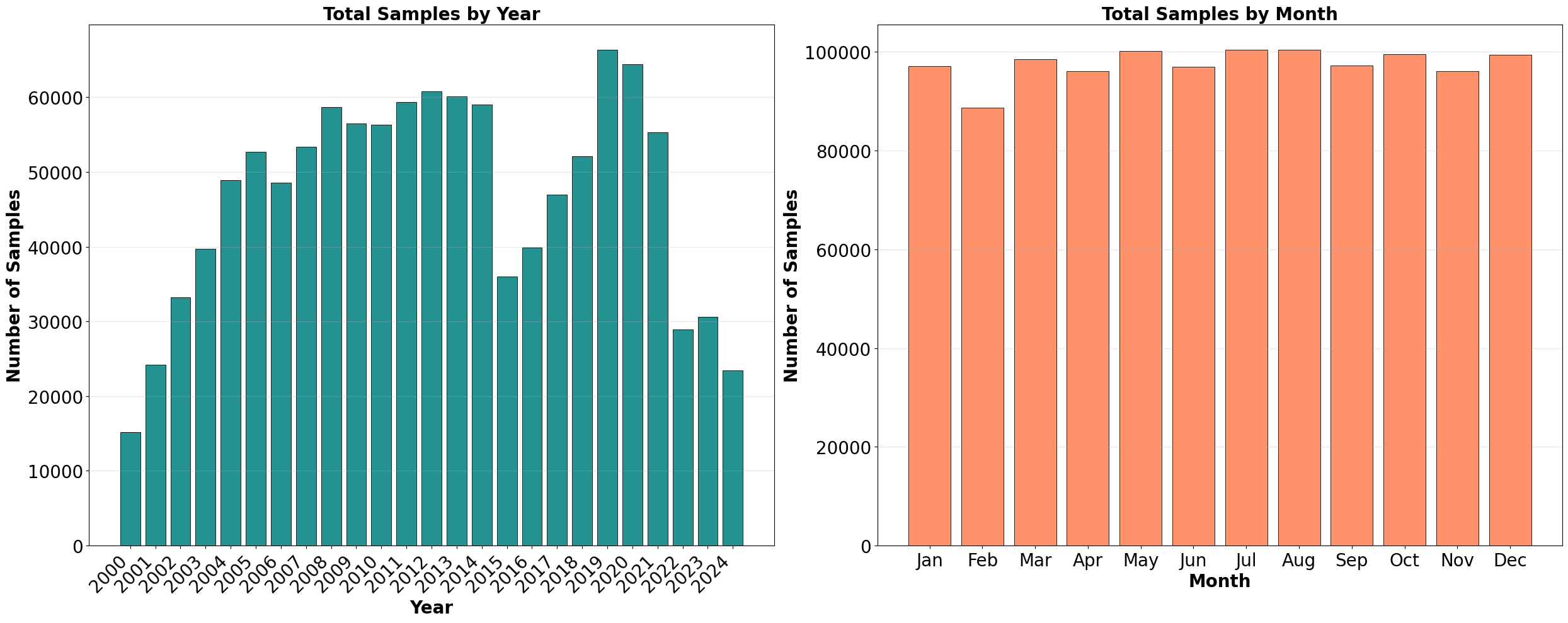}
    \vspace{-0.1cm}
    \caption{\small Temporal coverage of \textit{CarbonFluxBench}, showing the distribution of samples by year (left) and by month (right).}
    \vspace{-0.5cm}
    \label{fig:temp}
\end{figure}

Temporally, \textit{CarbonFluxBench} spans the period from 2000 to 2024, with the highest number of observations recorded in 2012 (Figure~\ref{fig:temp}). The monthly distribution of samples is ~uniform.

\begin{figure}
    \centering
    \includegraphics[width=1\linewidth]{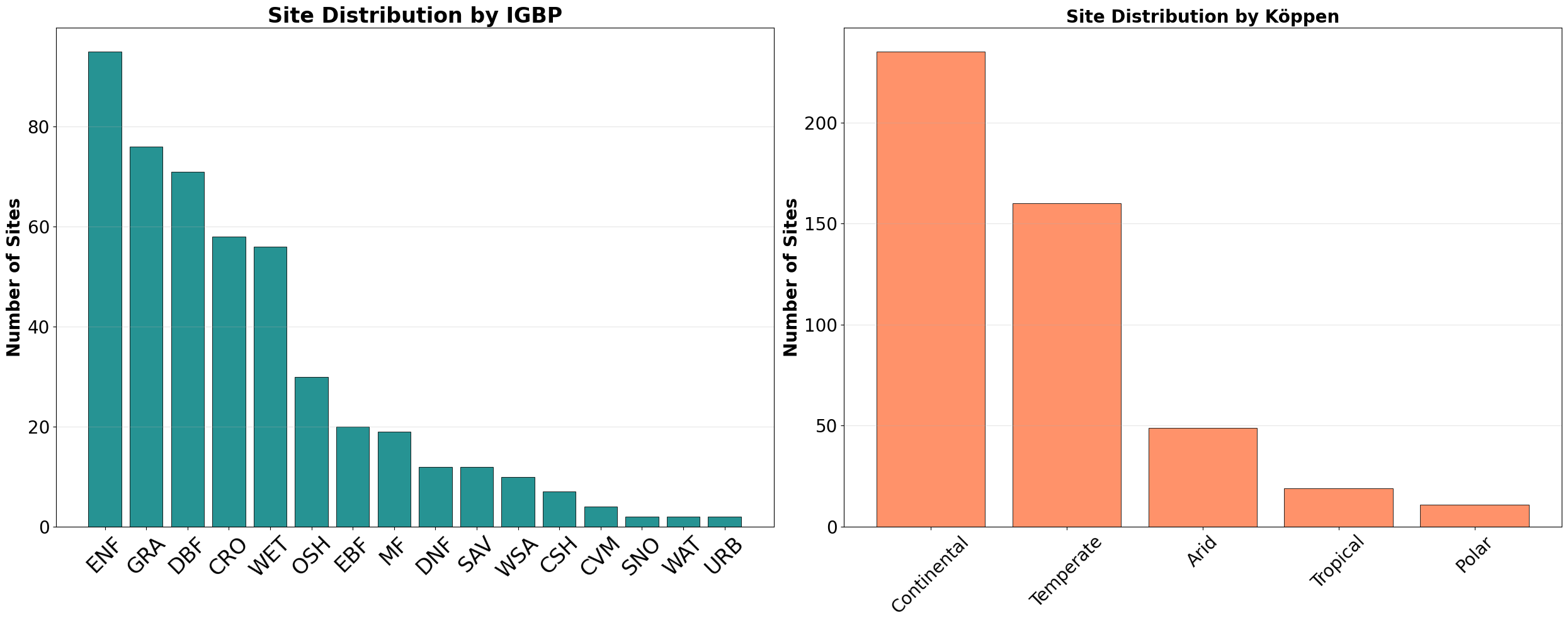}
    \vspace{-0.6cm}
    \caption{\small The distribution of site number by different IGBP (left) and Köppen (right) types, illustrating the large imbalance between geographical context of the available ground-true observations.}
    \vspace{-0.3cm}
    \label{fig:site_dist}
\end{figure}

To illustrate the pronounced imbalance in available EC observations -- one of the core challenges of our benchmark -- we present Figure ~\ref{fig:site_dist}, which shows the distribution of flux tower sites across IGBP vegetation types and K\"{o}ppen climate classes. The figure highlights the uneven representation of ecosystems and climates despite the apparent diversity of site locations.

\begin{figure}
    \centering
    \includegraphics[width=0.95\linewidth]{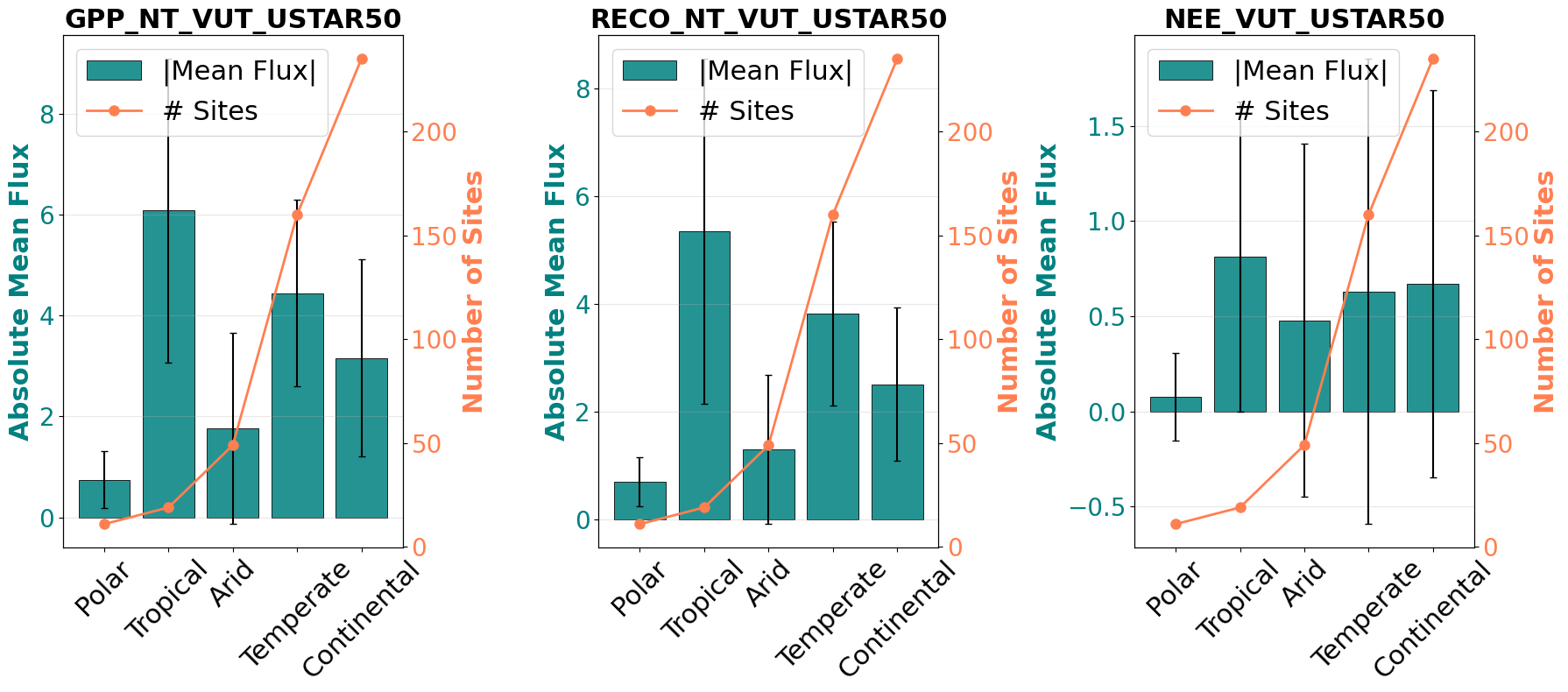}
    \vspace{-0.1cm}
    \caption{\small The average flux aggregates by Köppen climate type with bars representing standard deviation (left axis) and the number of site per climate type (right axis).}
    \vspace{-0.6cm}
    \label{fig:flux_magn}
\end{figure}

To extend this observation, Figure ~\ref{fig:flux_magn} reports the mean and standard deviation of each target variable aggregated by K\"{o}ppen climate class along with the number of sites per climate type. The figures show that tropical regions exhibit the largest absolute mean flux values across all targets, despite being the second least represented climate classes in the EC network.

%% file: eval.tex
\section{Experimental Design}

\subsection{Evaluation Metrics}

We evaluate model performance using three metrics: coefficient of determination (\textbf{R²}), root mean squared error (\textbf{RMSE}) in original flux units (gC m$^{-2}$ day$^{-1}$), and normalized mean absolute error (\textbf{nMAE}) computed as mean absolute error normalized by site-level mean flux magnitude to enable fair comparison across sites with substantially different flux ranges.
All metrics are computed per-site, then reported as quantiles (25th, 50th median, 75th percentiles) across the test site distribution. 

All test-set evaluations are performed exclusively on high-quality observations (NEE\_VUT\_USTAR50\_QC = 1) to ensure reliable assessment, while training utilizes the full quality spectrum via the quality-weighted loss function described in Section~\ref{sec:training}.

\subsection{Train-Test Split Strategy}
The dataset exhibits substantial imbalance across both IGBP vegetation types and Köppen climate classes, with some categories represented by as few as two sites. To comprehensively evaluate spatial generalization across diverse environmental contexts, we design two complementary stratified train-test splits: one stratified by ecosystem type (IGBP) and one by climate regime (Köppen). This dual evaluation protocol ensures that model performance is assessed separately across both ecological and climatic settings, revealing which dimension of environmental heterogeneity poses greater transfer learning challenges.

\begin{table}[h]
  \centering
  \caption{\small Train-test site partitions for IGBP vegetation (a) and Köppen climate (b) stratification schemes. All splits ensure representation of every category in both training and test sets despite severe class imbalance.}
  \label{tab:splits}
  \vspace{-0.5cm}
  \makebox[0.95\linewidth][c]{%
    \resizebox{1.5\linewidth}{!}{%
      \begin{tabular}{c}
        \begin{subtable}{\textwidth}
          \centering
          \caption{IGBP split}
          \label{tab:igbp}
          \begin{tabular}{lccc}
          \toprule
            IGBP Class & Raw & Train & Test \\
            \midrule
            Croplands (CRO) & 58 & 46 & 12 \\
            Closed Shrublands (CSH) & 7 & 4 & 3 \\
            Cropland/Natural Vegetation Mosaic (CVM) & 4 & 2 & 2 \\
            Deciduous Broadleaf Forests (DBF) & 71 & 57 & 14 \\
            Deciduous Needleleaf Forests (DNF) & 12 & 10 & 2 \\
            Evergreen Broadleaf Forests (EBF) & 20 & 16 & 4 \\
            Evergreen Needleleaf Forests (ENF) & 94 & 75 & 19 \\
            Grasslands (GRA) & 76 & 61 & 15 \\
            Mixed Forests (MF) & 18 & 14 & 4 \\
            Open Shrublands (OSH) & 30 & 24 & 6 \\
            Savannas (SAV) & 12 & 9 & 3 \\
            Snow and Ice (SNO) & 2 & 1 & 1 \\
            Urban and Built-up Lands (URB) & 2 & 1 & 1 \\
            Water Bodies (WAT) & 2 & 1 & 1 \\
            Permanent Wetlands (WET) & 56 & 45 & 11 \\
            Woody Savannas (WSA) & 10 & 5 & 5 \\
          \midrule
          \textbf{Total} & \textbf{474} & \textbf{371} & \textbf{103} \\
          \bottomrule
          \end{tabular}
        \end{subtable}
        \\[1em]
        \begin{subtable}{\textwidth}
          \centering
          \caption{Köppen climate split}
          \label{tab:koppen}
          \begin{tabular}{lccc}
          \toprule
          Köppen Class & Raw & Train & Test \\
          \midrule
          Tropical (A)     & 19  & 15  & 4  \\
          Arid (B)         & 49  & 39  & 10 \\
          Temperate (C)    & 160 & 128 & 32 \\
          Continental (D)  & 235 & 188 & 47 \\
          Polar (E)        & 11  & 9  & 2  \\
          \midrule
          \textbf{Total} & \textbf{474} & \textbf{379} & \textbf{95} \\
          \bottomrule
                \vspace{-1cm}
          \end{tabular}
        \end{subtable}
      \end{tabular}%
    }%
  }
\end{table}

Table~\ref{tab:splits} summarizes the resulting site partitions. For each stratification scheme, models are trained and evaluated independently, enabling direct comparison of generalization difficulty across vegetation types versus climate zones.

\textbf{IGBP-stratified split:} Several vegetation classes (SNO, URB, WAT) are represented by only two sites each. To ensure meaningful evaluation for these rare categories while retaining sufficient training signal, we apply a \textbf{0.5/0.5} train-test split for IGBP types with $\leq 10$ sites. For all other vegetation classes, we perform a standard \textbf{0.8/0.2} stratified split, yielding 371 training sites and 103 test sites.

\textbf{Köppen-stratified split:} Climate classes exhibit less severe imbalance than vegetation types, with the smallest class (Polar) containing 11 sites. We therefore apply a uniform \textbf{0.8/0.2} stratified split across all five climate zones, resulting in 379 training sites and 95 test sites.

By evaluating models under both splits, we can identify whether architectures generalize better across climate regimes, ecosystems, or struggle equally with both dimensions of spatial heterogeneity.

\subsection{Feature Sets}
In total, \textit{CarbonFluxBench} includes 165 features. Many of these covariates are redundant, and using the full feature set can be computationally expensive, particularly when combined with long temporal windows. To balance flexibility and efficiency, we define three nested ERA5-Land feature sets.

\begin{itemize}
    \item \textbf{Minimal (6 variables)}: A compact set capturing the primary biophysical drivers of terrestrial carbon exchange, based on domain knowledge and feature selections from prior carbon flux upscaling studies \cite{nathaniel2023metaflux, renganathan2025task, rozanov2025tamrl_carbon}: 2-meter air temperature, total precipitation, surface net solar radiation, total evaporation, and leaf area indices for high and low vegetation. This configuration prioritizes the most direct controls on photosynthesis and respiration.
    \item \textbf{Standard (36)}: an extended set that augments the minimal configuration with wind, air pressure, more detailed radiation, and soil variables.
    \item \textbf{Full (150)}: the complete set of all ERA5-Land variables available via GEE.
\end{itemize}

We report baseline results produced on the \textbf{Minimal} feature set in Section~\ref{sec:results} as it is the most constrained and computationally efficient setting, whereas two other splits are reported in Appendix~\ref{app:baselines}.

\vspace{-0.2cm}
\subsection{Baseline Models}

We evaluate two categories of baseline architectures: \textbf{static} models that operate on single time-step feature vectors, and \textbf{temporal} models that explicitly leverage sequential dependencies. All neural models are trained in a multi-task setting to jointly predict GPP, RECO, and NEE.
Static baselines operate on feature vectors of shape $(1, F)$, where $F$ denotes the predictors for a single day. This category includes widely-used tree-based methods, such as XGBoost \cite{chen2016xgboost} and LightGBM \cite{ke2017lightgbm}, which dominate carbon flux upscaling literature.

Temporal baselines capture sequential dependencies using sliding windows of $W = 30$ consecutive days with a stride of 15 days. We evaluate recurrent architectures including LSTM \cite{hochreiter1997long} and GRU \cite{cho2014learning}, along with their concatenated-input variants CT-LSTM and CT-GRU, where categorical variables (IGBP type, Köppen class) are one-hot encoded and concatenated at each time step. We additionally consider transformer-based architectures: an encoder-only Transformer \cite{vaswani2017attention} and a Patch-Transformer based on PatchTST \cite{nie2022time}. Finally, we include TAM-RL \cite{renganathan2025task}, a transfer-learning-specific architecture designed for cross-domain generalization in environmental applications.

\vspace{-0.2cm}
\subsection{Training Protocol}\label{sec:training}

All neural network models are trained using a composite loss function incorporating quality weighting, class balancing, and physics-based flux balance constraints, as described in prior work \cite{rozanov2025tamrl_carbon}. Due to the algorithmic limitations of tree-based methods, their loss function is similar excluding the flux balance component which requires the model to be multitask. Hyperparameter optimization is performed via 5-fold cross-validation stratified by Köppen climate class, and final predictions are obtained by averaging across 10 models trained with different random seeds. Full details on the loss function formulation and hyperparameter configurations are provided in Appendix~\ref{app:training}.

%% file: results.tex
\vspace{-0.1cm}
\section{Results} \label{sec:results}

\begin{table*}
\centering
\resizebox{0.87\textwidth}{!}{
\begin{minipage}{\textwidth}
\caption{Baseline zero-shot performance on the \textit{Minimal} feature set of \textit{CarbonFluxBench}. Results show the median over an ensemble of 10 models trained with different random seeds, with superscripts and subscripts denoting the 75th and 25th percentiles, respectively. Arrows indicate the direction of improvement.}
\vspace{-0.2cm}
\label{tab:zero-baselines}
\centering
\begin{tabular}{ll ccc ccc ccc}
\toprule
 &  & \multicolumn{9}{c}{\textbf{Targets}} \\
\cmidrule(lr){3-11}
\textbf{Split} & \textbf{Model}
 & \multicolumn{3}{c}{\textbf{GPP}}
 & \multicolumn{3}{c}{\textbf{RECO}}
 & \multicolumn{3}{c}{\textbf{NEE}} \\
\cmidrule(lr){3-5} \cmidrule(lr){6-8} \cmidrule(lr){9-11}
 & 
 & $R^2\uparrow$ & RMSE$\downarrow$ & nMAE$\downarrow$
 & $R^2\uparrow$ & RMSE$\downarrow$ & nMAE$\downarrow$
 & $R^2\uparrow$ & RMSE$\downarrow$ & nMAE$\downarrow$ \\
\midrule
\multirow{9}{*}{IGBP}
 & XGBoost & $0.501^{0.700}_{-0.112}$ & $1.987^{2.758}_{1.450}$ & $0.463^{0.817}_{0.349}$ & $0.060^{0.574}_{-1.308}$ & $1.553^{2.450}_{1.044}$ & $0.432^{0.959}_{0.308}$ & $0.129^{0.381}_{-0.508}$ & $1.694^{2.215}_{1.076}$ & $2.303^{5.259}_{1.558}$ \\[0.5ex]
 & LightGBM & $0.540^{0.731}_{0.028}$ & $1.797^{2.431}_{1.165}$ & $0.441^{0.692}_{0.339}$ & $0.350^{0.664}_{-0.509}$ & $1.223^{1.870}_{0.841}$ & $0.385^{0.528}_{0.290}$ & $\textbf{0.219}^{0.404}_{-0.162}$ & $1.492^{2.079}_{0.855}$ & $2.130^{4.023}_{1.299}$ \\[0.5ex]
 & LSTM & $0.547^{0.747}_{0.019}$ & $1.959^{2.697}_{1.159}$ & $0.469^{0.690}_{0.351}$ & $0.389^{0.616}_{-0.411}$ & $1.280^{2.078}_{0.816}$ & $0.399^{0.579}_{0.297}$ & $0.170^{0.405}_{-0.192}$ & $1.493^{2.074}_{0.962}$ & $2.200^{4.347}_{1.351}$ \\[0.5ex]
 & CT-LSTM & $0.573^{0.743}_{0.179}$ & $\textbf{1.593}^{2.279}_{1.049}$ & $\textbf{0.437}^{0.642}_{0.315}$ & $\textbf{0.451}^{0.643}_{-0.241}$ & $1.260^{1.776}_{0.734}$ & $0.369^{0.562}_{0.290}$ & $0.141^{0.387}_{-0.092}$ & $1.494^{2.130}_{0.882}$ & $2.188^{4.169}_{1.359}$ \\[0.5ex]
 & GRU & $0.438^{0.640}_{0.077}$ & $1.804^{2.763}_{1.298}$ & $0.514^{0.689}_{0.387}$ & $0.211^{0.524}_{-0.204}$ & $1.372^{1.851}_{0.961}$ & $0.433^{0.568}_{0.321}$ & $0.201^{0.358}_{-0.073}$ & $1.602^{2.083}_{0.881}$ & $2.163^{4.049}_{1.355}$ \\[0.5ex]
 & CT-GRU & $0.529^{0.732}_{0.094}$ & $1.643^{2.380}_{1.170}$ & $0.452^{0.695}_{0.341}$ & $0.409^{0.624}_{-0.246}$ & $1.231^{1.849}_{0.789}$ & $0.397^{0.574}_{0.292}$ & $0.137^{0.382}_{-0.143}$ & $1.461^{2.173}_{0.960}$ & $2.205^{3.963}_{1.326}$ \\[0.5ex]
 & Transformer & $0.526^{0.736}_{0.221}$ & $1.704^{2.321}_{1.118}$ & $0.452^{0.672}_{0.313}$ & $0.385^{0.657}_{-0.242}$ & $\textbf{1.214}^{1.772}_{0.891}$ & $0.378^{0.595}_{0.289}$ & $0.193^{0.419}_{0.006}$ & $\textbf{1.384}^{2.093}_{0.832}$ & $\textbf{2.124}^{3.963}_{1.279}$ \\[0.5ex]
 & Patch-Transformer & $0.435^{0.690}_{0.093}$ & $1.830^{2.646}_{1.271}$ & $0.494^{0.684}_{0.356}$ & $0.374^{0.593}_{-0.103}$ & $1.298^{1.887}_{0.953}$ & $0.390^{0.537}_{0.322}$ & $0.118^{0.334}_{-0.209}$ & $1.570^{2.084}_{0.928}$ & $2.293^{4.120}_{1.349}$ \\[0.5ex]
 & TAM-RL & $\textbf{0.574}^{0.758}_{0.251}$ & $1.610^{2.355}_{1.080}$ & $0.459^{0.656}_{0.297}$ & $0.393^{0.662}_{-0.248}$ & $1.242^{1.842}_{0.819}$ & $\textbf{0.368}^{0.562}_{0.290}$ & $0.204^{0.429}_{-0.101}$ & $1.475^{2.013}_{0.894}$ & $2.140^{4.109}_{1.311}$ \\[0.5ex]
\midrule
\multirow{9}{*}{K\"{o}ppen}
 & XGBoost & $0.469^{0.689}_{0.139}$ & $2.069^{2.556}_{1.376}$ & $0.553^{0.668}_{0.385}$ & $0.426^{0.633}_{-0.177}$ & $1.298^{2.166}_{0.943}$ & $0.390^{0.612}_{0.304}$ & $0.162^{0.379}_{-0.577}$ & $1.619^{2.114}_{1.214}$ & $2.654^{6.867}_{1.274}$ \\[0.5ex]
 & LightGBM & $0.608^{0.764}_{0.286}$ & $1.717^{2.354}_{1.137}$ & $0.443^{0.626}_{0.320}$ & $0.541^{0.700}_{0.186}$ & $1.168^{1.649}_{0.812}$ & $\textbf{0.343}^{0.478}_{0.275}$ & $0.283^{0.497}_{0.052}$ & $1.333^{1.926}_{0.936}$ & $\textbf{2.046}^{5.359}_{1.224}$ \\[0.5ex]
 & LSTM & $0.608^{0.758}_{0.231}$ & $1.669^{2.310}_{1.189}$ & $0.436^{0.614}_{0.333}$ & $0.401^{0.717}_{-0.150}$ & $1.162^{1.867}_{0.782}$ & $0.366^{0.545}_{0.264}$ & $0.241^{0.477}_{-0.058}$ & $\textbf{1.296}^{2.078}_{0.977}$ & $2.330^{5.349}_{1.085}$ \\[0.5ex]
 & CT-LSTM & $\textbf{0.648}^{0.772}_{0.241}$ & $\textbf{1.674}^{2.418}_{1.119}$ & $\textbf{0.397}^{0.610}_{0.305}$ & $0.489^{0.698}_{0.016}$ & $1.185^{1.872}_{0.822}$ & $0.364^{0.539}_{0.265}$ & $\textbf{0.296}^{0.479}_{-0.052}$ & $1.364^{1.911}_{0.936}$ & $2.226^{5.405}_{1.229}$ \\[0.5ex]
 & GRU & $0.556^{0.743}_{0.264}$ & $1.778^{2.387}_{1.212}$ & $0.445^{0.628}_{0.339}$ & $0.387^{0.640}_{-0.170}$ & $1.281^{1.775}_{0.889}$ & $0.357^{0.575}_{0.292}$ & $0.178^{0.451}_{-0.072}$ & $1.317^{2.056}_{0.969}$ & $2.238^{5.404}_{0.998}$ \\[0.5ex]
 & CT-GRU & $0.564^{0.730}_{0.175}$ & $1.827^{2.296}_{1.298}$ & $0.469^{0.682}_{0.343}$ & $0.475^{0.657}_{-0.203}$ & $1.285^{1.799}_{0.897}$ & $0.373^{0.526}_{0.282}$ & $0.159^{0.406}_{-0.129}$ & $1.422^{2.106}_{1.039}$ & $2.355^{5.697}_{1.129}$ \\[0.5ex]
 & Transformer & $0.600^{0.767}_{0.143}$ & $1.756^{2.382}_{1.183}$ & $0.429^{0.656}_{0.310}$ & $0.485^{0.694}_{-0.003}$ & $1.153^{1.720}_{0.951}$ & $0.367^{0.522}_{0.266}$ & $0.212^{0.464}_{-0.056}$ & $1.362^{1.981}_{1.031}$ & $2.295^{5.986}_{1.108}$ \\[0.5ex]
 & Patch-Transformer & $0.506^{0.694}_{0.123}$ & $2.003^{2.637}_{1.266}$ & $0.488^{0.693}_{0.363}$ & $0.527^{0.679}_{-0.046}$ & $1.255^{1.891}_{0.936}$ & $0.380^{0.540}_{0.293}$ & $0.107^{0.359}_{-0.046}$ & $1.500^{2.077}_{1.024}$ & $2.314^{5.490}_{1.254}$ \\[0.5ex]
 & TAM-RL & $0.644^{0.762}_{0.224}$ & $\textbf{1.674}^{2.413}_{1.113}$ & $0.444^{0.628}_{0.308}$ & $\textbf{0.593}^{0.725}_{0.053}$ & $\textbf{1.125}^{1.747}_{0.848}$ & $0.356^{0.504}_{0.262}$ & $0.270^{0.468}_{0.057}$ & $1.341^{1.874}_{0.934}$ & $2.277^{5.657}_{1.036}$ \\[0.5ex]
\bottomrule
      \vspace{-0.5cm}
\end{tabular}

\end{minipage}
}
\end{table*}

\subsection{Zero-Shot Performance}

Table~\ref{tab:zero-baselines} presents baseline performance on the Minimal feature set with results reported as median site-level performance and interquartile range across test sites (subscript = 25th percentile, superscript = 75th percentile). Temporal architectures lead on GPP and RECO across both stratification schemes: under IGBP, TAM-RL and CT-LSTM achieve equivalent median GPP ($R^2 = 0.574$ and $0.573$); under K\"{o}ppen, CT-LSTM leads GPP ($R^2 = 0.648$) and TAM-RL leads RECO ($R^2 = 0.593$). Performance among tree-based baselines diverges sharply: LightGBM is competitive with temporal models on GPP under both splits (IGBP: $0.540$, K\"{o}ppen: $0.608$), while XGBoost trails substantially (IGBP GPP: $0.501$, K\"{o}ppen RECO: $0.426$). 

NEE behaves structurally differently than GPP and RECO. LightGBM, which is trained with only a single day of temporal context, achieves the best median NEE under IGBP ($R^2 = 0.219$) and remains competitive under K\"{o}ppen ($R^2 = 0.283$), where CT-LSTM is the strongest temporal model ($R^2 = 0.296$). The lack of clear advantage from sequence modeling on NEE might stem from the fact that this variable is a small residual between two larger opposing fluxes (GPP and RECO). As a result, neural models may overfit to noise rather than signal. Median nMAE for NEE ranges from $2.046$ to $2.654$, roughly $5\text{--}6\times$ higher than GPP nMAE ($0.397\text{--}0.553$). This pattern, together with broader findings in the carbon flux literature \cite{tramontana2016predicting}, motivates treating NEE as a distinct modeling problem rather than a third instance of the GPP/RECO task.

Worst-case site performance (25th percentile) discriminates models more sharply than the median. Under IGBP GPP, TAM-RL maintains $R^2 = 0.251$ at the 25th percentile, exceeding Transformer ($0.221$), CT-LSTM ($0.179$), and LSTM ($0.019$). Under K\"{o}ppen RECO, TAM-RL is the only temporal model with a positive 25th percentile ($R^2 = 0.053$); LSTM ($-0.150$), Transformer ($-0.003$), and Patch-Transformer ($-0.046$) all degrade below the mean predictor on the worst quartile. LightGBM is again notable here: its 25th-percentile K\"{o}ppen RECO ($R^2 = 0.186$) exceeds every temporal baseline, suggesting that tree-based architectures are competitive in this task.

\vspace{-0.25cm}
\subsection{Vegetation vs. Climate Stratification}

The two split schemes expose complementary failure modes. K\"{o}ppen yields higher median performance than IGBP for the strongest models on GPP and RECO, suggesting that climate-based test sites are on average closer to training-distribution dynamics than vegetation-type holdouts. However, K\"{o}ppen also produces wider site-level interquartile ranges for NEE (CT-LSTM K\"{o}ppen NEE IQR $= 0.531$ vs. IGBP $= 0.479$), separating sites more sharply into well-transferred and failing groups than IGBP. This bimodal K\"{o}ppen behavior is clearest at the 25th percentile. Several models with strong median K\"{o}ppen RECO collapse on the lower quartile (Transformer: median $0.485$, 25th $-0.003$; LSTM: median $0.401$, 25th $-0.150$). TAM-RL is the only temporal architecture that maintains positive 25th-percentile GPP, RECO, NEE under K\"{o}ppen, and its 25th-percentile GPP under IGBP ($0.251$) likewise leads, indicating that its task-aware nature allows more uniform generalization.

\subsection{Process-based baseline} In this work, we do not attempt a direct comparison with process-based models. Yet as an additional baseline, we evaluate the MiCASA model \cite{van2010global}, a NASA-distributed process-based biogeochemical model. Specifically, we compare net primary production (NPP), heterotrophic ecosystem respiration ($R_h$), and NEE against EC observations. Due to spatial gaps in the MiCASA product at some flux tower locations, we cannot directly compare it against the data-driven benchmarks in Table~\ref{tab:zero-baselines}. However, after discarding missing samples ($\approx$1.2\%), we compute performance metrics on the overlapping subset (Table~\ref{tab:micasa-comparison}).

The process-based model exhibits substantially lower performance than all data-driven baselines. However, two caveats limit direct interpretation: (1) MiCASA operates at $0.1^{\circ}$ spatial resolution, introducing scale mismatch with site-level EC footprints and our remote sensing data, and (2) MiCASA predicts NPP, not GPP, which are related quantities that differ by autotrophic respiration of a plant. Despite these limitations, the comparison demonstrates that even simple tree-based models substantially outperform process-based systems for spatial carbon flux prediction, motivating the benchmark's focus on advancing ML approaches.

\begin{table}[h]
\centering
\vspace{-0.1cm}
\caption{\small Performance of the process-based MiCASA model on \textit{CarbonFluxBench}. Results show median site-level metrics across test sites. MiCASA predicts NPP (not GPP) and $R_h$ (not RECO), which are related but distinct physical quantities.}
\vspace{-0.4cm}
\label{tab:micasa-comparison}
\resizebox{0.7\columnwidth}{!}{
\begin{tabular}{llcccc}
\toprule
\textbf{Split} & \textbf{Target} & $R^2\uparrow$ & RMSE$\downarrow$ & nMAE$\downarrow$ \\
\midrule
\multirow{3}{*}{K\"{o}ppen} 
 & GPP & 0.124 & 2.958 & 0.806 \\
 & RECO & $-$0.174 & 2.303 & 0.960 \\
 & NEE & 0.133 & 1.733 & 0.891 \\
\midrule
\multirow{3}{*}{IGBP} 
 & GPP & 0.114 & 2.920 & 0.797 \\
 & RECO & $-$0.228 & 2.212 & 0.985 \\
 & NEE & $-$0.005 & 1.750 & 0.972 \\
\bottomrule
\end{tabular}
}
\vspace{-0.5cm}
\end{table}

%% file: discussion.tex
\section{Discussions}

Beyond serving as a benchmark, \textit{CarbonFluxBench} is designed as a testbed for scientific discovery. Models developed and evaluated within this framework have the potential to improve operational carbon flux products. At the same time, systematic analysis of where and why models succeed or fail in spatial transfer can yield insights that inform both ML research and the design of future ecological field campaigns. Below, we outline a set of key research opportunities enabled by \textit{CarbonFluxBench}:

\textit{(1) Feature engineering and selection:} While we provide baseline scores, we mainly focus on the Minimal set. Therefore, systematic exploration of Standard and Full feature sets is left to future users of the benchmark.
\textit{(2) Transfer learning:} Except for TAM-RL, our baselines use standard supervised learning. Can adversarial domain adaptation, domain-invariant representations, or multi-source transfer learning improve zero-shot generalization? Do they perform better on spatial transfer compared to classical supervised methods? \textit{(3) Uncertainty quantification (UQ):} In \textit{CarbonFluxBench}, we intentionally use ensembles of baselines trained with different seeds to account for model uncertainty. However, broader questions of UQ should be answered. \textit{(4) Task heterogeneity and negative transfer:} Baselines jointly predict GPP, RECO, and NEE, but it remains unclear whether sharing a backbone across these structurally different targets introduces negative transfer. GPP and RECO are opposing fluxes of comparable magnitude while NEE is their small noisy residual, suggesting that target-specific architectures or mixture-of-experts heads may outperform a shared model.
\textit{(5) Ecosystem and climate heterogeneity:} Beyond task heterogeneity, IGBP and Köppen stratifications expose distinct axes of domain heterogeneity: vegetation structure versus climate regime. This direction asks whether ecosystem- or climate-specific model components can reduce these failures, and how task and domain heterogeneity interact, remains an open question.  \textit{(6) Stratified and biome-specific evaluation:}  While the previous question asks where models fail, future work should thoroughly investigate model performance by individual IGBP and Köppen classes. This research may reveal which ecosystems pose the greatest transfer challenges and be used to identify where improved models or targeted data collection are most needed. \textit{(7) Self-supervised learning (SSL):} Pretraining on unlabeled satellite time and meteorological series could leverage the vast spatial-temporal coverage of gridded data beyond the limited labeled flux tower sites. \textit{(8) Knowledge-guided ML:} The existing physical models are ideal candidates for pre-training of the ML models with later fine-tuning on the ground-true EC observations. This direction can be combined with the ideas of likelihood inference, surrogate modeling, and physics-informed priors.

%% file: conclusion.tex
\vspace{-0.1cm}
\section{Conclusion}

We introduce \textit{CarbonFluxBench}, the first benchmark for spatial transfer learning in carbon flux upscaling, providing 474 eddy covariance sites with satellite and meteorological features, dual stratified evaluation protocols, and reproducible baseline implementations. Our zero-shot evaluation reveals that temporal models outperform static baselines, with TAM-RL achieving not only competitive median performance but critically improved robustness through consistently higher 25th percentile metrics, indicating fewer catastrophic failures at challenging sites. This demonstrates that architectures explicitly designed for cross-domain generalization can improve worst-case spatial transfer—essential for reliable carbon monitoring across all ecosystems. However, all models struggle with NEE prediction and exhibit severe degradation on underrepresented biomes, with climate-stratified evaluation proving more challenging for robust generalization.

Beyond carbon science, \textit{CarbonFluxBench} addresses a critical gap in ML benchmarking: spatial transfer learning for time series regression. By providing quantile-based evaluation that exposes tail performance rather than masking failures through averaging, \textit{CarbonFluxBench} enables rigorous assessment of spatial transfer. 

%% file: appendix_a.tex
\section{Baseline results for the \textit{Standard} and \textit{Full} feature sets}
\label{app:baselines}

To validate the broader feature architecture of \textit{CarbonFluxBench}, we trained all recurrent baselines on the \textit{Standard} (36 variables) and \textit{Full} (150 variables) feature sets using 10-member ensembles (we ommit tree-based baselines due to compute-time on large feature sets). Results are presented in Tables~\ref{tab:standard-baselines} and~\ref{tab:full-baselines}, with the \textit{Minimal} set (Table~\ref{tab:zero-baselines}) as reference.

According to our experiments, \textbf{GPP and RECO show mild degradation with feature richness, but NEE improves.} For GPP, median performance degrades modestly as feature dimensionality grows. Under IGBP, CT-LSTM GPP $R^2$ drops from $0.573$ (Minimal) to $0.551$ (Standard) and $0.561$ (Full); TAM-RL follows a similar pattern ($0.574 \rightarrow 0.539 \rightarrow 0.548$). Under K\"{o}ppen, the degradation is more pronounced: CT-LSTM GPP falls from $0.648$ to $0.625$ to $0.584$. RECO behavior is mixed, often peaking on the Standard set (e.g., IGBP CT-LSTM RECO: $0.451 \rightarrow 0.475 \rightarrow 0.443$).

In contrast, some models \textit{benefit} from richer features on the NEE prediction task with the largest gains seen on the Standard set. Under IGBP, LSTM NEE improves from $R^2 = 0.170$ (Minimal) to $0.224$ (Standard), and CT-LSTM improves from $0.141$ to $0.171$ to $0.207$. Under K\"{o}ppen, LSTM NEE jumps from $0.241$ to $0.315$ on Standard. This divergent behavior across targets reinforces the hypothesis that NEE is a structurally distinct prediction problem: as a small residual between GPP and RECO, NEE may benefit from auxiliary state variables (soil moisture, soil temperature, additional radiation components) that help disentangle the two opposing fluxes, while GPP and RECO are already well-described by the Minimal driver set.

\textbf{Tail robustness degrades despite stable medians.} For GPP under IGBP, the 25th percentile $R^2$ degrades sharply across feature sets: TAM-RL falls from $0.251$ (Minimal) to $0.099$ (Standard) and $0.087$ (Full); CT-LSTM degrades from $0.179$ to $0.099$ to $0.111$. This indicates that while typical sites remain well-predicted, additional features increase the risk of catastrophic failure on out-of-distribution sites. The gap between median and 25th-percentile performance widens with feature dimensionality, suggesting that feature-rich models overfit to training-site-specific patterns rather than learning transferable representations.

\textbf{Architectural ranking partially shifts across feature sets.} CT-LSTM remains the strongest GPP model under both stratification schemes across all feature sets, and TAM-RL retains its RECO advantage under K\"{o}ppen. However, NEE ranking shifts: the simplest temporal model (LSTM) emerges as best on the Standard set under both stratifications, suggesting that more complex architectures may overfit when given richer features for an inherently noisy target. Patch-Transformer consistently underperforms across configurations. Practitioners should select architectures on a per-target basis rather than assuming a single model dominates across all fluxes.

Despite these findings, the Standard and Full sets enable investigation of several research directions:

\begin{enumerate}
    \item \textbf{Feature selection and regularization:} Future work could identify which of the 150 variables actually improve OOD generalization.
    \item \textbf{Pre-training and representation learning:} Self-supervised pre-training on the Full set (masked autoencoding, contrastive learning across sites) may extract representations that supervised training alone cannot, particularly given the NEE-specific improvement observed with richer features.
    \item \textbf{Domain-specific fine-tuning:} While zero-shot GPP transfer degrades, few-shot adaptation might benefit from rich features at test sites. A model could meta-learn on the Full set during training, then leverage site-specific features (soil properties, local climate) when small numbers of labeled observations become available.
    \item \textbf{Target-specific architectures:} The divergent feature-richness response between NEE and GPP/RECO suggests that single-target models or architectures with target-specific feature pathways may outperform shared backbones.
\end{enumerate}

\begin{table*}[t]
\centering
\resizebox{0.87\textwidth}{!}{
\begin{minipage}{\textwidth}
\caption{Baseline zero-shot performance on the \textit{Standard} feature set of \textit{CarbonFluxBench}. Results show the median over an ensemble of 10 models trained with different random seeds, with superscripts and subscripts denoting the 75th and 25th percentiles, respectively. Arrows indicate the direction of improvement.}
\vspace{-0.2cm}
\label{tab:standard-baselines}
\centering
\begin{tabular}{ll ccc ccc ccc}
\toprule
 &  & \multicolumn{9}{c}{\textbf{Targets}} \\
\cmidrule(lr){3-11}
\textbf{Split} & \textbf{Model}
 & \multicolumn{3}{c}{\textbf{GPP}}
 & \multicolumn{3}{c}{\textbf{RECO}}
 & \multicolumn{3}{c}{\textbf{NEE}} \\
\cmidrule(lr){3-5} \cmidrule(lr){6-8} \cmidrule(lr){9-11}
 & 
 & $R^2\uparrow$ & RMSE$\downarrow$ & nMAE$\downarrow$
 & $R^2\uparrow$ & RMSE$\downarrow$ & nMAE$\downarrow$
 & $R^2\uparrow$ & RMSE$\downarrow$ & nMAE$\downarrow$ \\
\midrule
\multirow{7}{*}{IGBP}
 & LSTM & $0.449^{0.724}_{-0.025}$ & $1.799^{2.524}_{1.245}$ & $0.490^{0.619}_{0.352}$ & $0.361^{0.693}_{-0.437}$ & $1.320^{1.876}_{0.765}$ & $0.403^{0.534}_{0.281}$ & $\textbf{0.224}^{0.432}_{-0.157}$ & $1.457^{2.063}_{0.868}$ & $2.153^{4.071}_{1.306}$ \\[0.5ex]
 & CT-LSTM & $\textbf{0.551}^{0.759}_{0.099}$ & $\textbf{1.646}^{2.203}_{1.144}$ & $0.446^{0.639}_{0.315}$ & $\textbf{0.475}^{0.683}_{-0.134}$ & $1.174^{1.730}_{0.761}$ & $\textbf{0.349}^{0.510}_{0.276}$ & $0.171^{0.417}_{-0.180}$ & $1.454^{2.025}_{0.858}$ & $2.051^{3.809}_{1.314}$ \\[0.5ex]
 & GRU & $0.452^{0.677}_{0.111}$ & $1.784^{2.791}_{1.287}$ & $0.522^{0.661}_{0.386}$ & $0.325^{0.596}_{-0.229}$ & $1.296^{1.905}_{0.887}$ & $0.416^{0.548}_{0.315}$ & $0.095^{0.378}_{-0.131}$ & $1.543^{2.084}_{0.977}$ & $2.179^{4.245}_{1.451}$ \\[0.5ex]
 & CT-GRU & $0.506^{0.710}_{0.077}$ & $1.731^{2.423}_{1.178}$ & $0.477^{0.690}_{0.338}$ & $0.419^{0.645}_{-0.171}$ & $1.234^{1.871}_{0.786}$ & $0.382^{0.505}_{0.294}$ & $0.176^{0.368}_{-0.388}$ & $1.527^{2.166}_{0.995}$ & $2.111^{5.093}_{1.311}$ \\[0.5ex]
 & Transformer & $0.517^{0.731}_{0.184}$ & $1.673^{2.303}_{1.152}$ & $\textbf{0.430}^{0.656}_{0.327}$ & $0.429^{0.682}_{-0.139}$ & $1.204^{1.780}_{0.805}$ & $0.360^{0.526}_{0.278}$ & $0.192^{0.434}_{-0.080}$ & $1.481^{1.988}_{0.867}$ & $1.991^{3.822}_{1.345}$ \\[0.5ex]
 & Patch-Transformer & $0.426^{0.695}_{0.102}$ & $1.759^{2.455}_{1.215}$ & $0.471^{0.668}_{0.357}$ & $0.392^{0.614}_{-0.220}$ & $1.215^{1.854}_{0.891}$ & $0.375^{0.535}_{0.313}$ & $0.130^{0.304}_{-0.245}$ & $1.578^{2.128}_{0.904}$ & $2.188^{3.903}_{1.403}$ \\[0.5ex]
 & TAM-RL & $0.539^{0.748}_{0.099}$ & $1.684^{2.375}_{1.110}$ & $0.453^{0.648}_{0.317}$ & $0.445^{0.690}_{-0.118}$ & $\textbf{1.126}^{1.707}_{0.757}$ & $0.356^{0.542}_{0.285}$ & $0.211^{0.458}_{-0.103}$ & $\textbf{1.423}^{2.041}_{0.833}$ & $\textbf{1.930}^{3.676}_{1.266}$ \\[0.5ex]
\midrule
\multirow{7}{*}{K\"{o}ppen}
 & LSTM & $0.577^{0.741}_{0.230}$ & $1.843^{2.268}_{1.273}$ & $0.456^{0.623}_{0.327}$ & $0.428^{0.681}_{0.002}$ & $1.201^{1.821}_{0.874}$ & $0.364^{0.519}_{0.278}$ & $\textbf{0.315}^{0.468}_{-0.036}$ & $1.365^{2.035}_{0.879}$ & $2.120^{5.534}_{1.145}$ \\[0.5ex]
 & CT-LSTM & $\textbf{0.625}^{0.750}_{0.142}$ & $\textbf{1.712}^{2.585}_{1.151}$ & $\textbf{0.421}^{0.600}_{0.309}$ & $0.518^{0.698}_{0.046}$ & $1.232^{1.786}_{0.869}$ & $\textbf{0.358}^{0.531}_{0.277}$ & $0.302^{0.485}_{0.028}$ & $1.377^{1.876}_{0.935}$ & $2.194^{5.231}_{1.224}$ \\[0.5ex]
 & GRU & $0.530^{0.729}_{0.105}$ & $1.820^{2.549}_{1.271}$ & $0.465^{0.643}_{0.347}$ & $0.386^{0.673}_{-0.212}$ & $1.309^{1.754}_{0.880}$ & $0.371^{0.587}_{0.296}$ & $0.210^{0.414}_{-0.184}$ & $1.405^{2.106}_{0.950}$ & $2.198^{5.490}_{1.186}$ \\[0.5ex]
 & CT-GRU & $0.534^{0.736}_{0.158}$ & $1.820^{2.415}_{1.314}$ & $0.459^{0.676}_{0.345}$ & $0.495^{0.686}_{-0.014}$ & $1.193^{1.739}_{0.820}$ & $0.391^{0.554}_{0.288}$ & $0.131^{0.440}_{-0.229}$ & $1.401^{2.078}_{1.058}$ & $2.510^{5.810}_{1.246}$ \\[0.5ex]
 & Transformer & $0.582^{0.746}_{0.182}$ & $1.866^{2.395}_{1.162}$ & $0.438^{0.628}_{0.319}$ & $0.486^{0.695}_{0.001}$ & $1.220^{1.861}_{0.904}$ & $0.359^{0.530}_{0.284}$ & $0.247^{0.481}_{0.022}$ & $\textbf{1.351}^{1.818}_{1.002}$ & $2.216^{5.434}_{1.152}$ \\[0.5ex]
 & Patch-Transformer & $0.486^{0.695}_{0.153}$ & $1.992^{2.633}_{1.342}$ & $0.487^{0.678}_{0.346}$ & $0.477^{0.685}_{-0.085}$ & $1.271^{1.741}_{0.903}$ & $0.380^{0.536}_{0.290}$ & $0.143^{0.320}_{-0.109}$ & $1.521^{2.103}_{0.956}$ & $2.255^{5.807}_{1.297}$ \\[0.5ex]
 & TAM-RL & $0.601^{0.744}_{0.222}$ & $1.749^{2.272}_{1.190}$ & $0.451^{0.645}_{0.312}$ & $\textbf{0.530}^{0.680}_{-0.000}$ & $\textbf{1.166}^{1.678}_{0.869}$ & $0.359^{0.520}_{0.264}$ & $0.249^{0.484}_{0.008}$ & $1.357^{1.924}_{0.897}$ & $\textbf{2.070}^{5.526}_{1.126}$ \\[0.5ex]
\bottomrule
      \vspace{-0.7cm}
\end{tabular}

\end{minipage}
}
\end{table*}
\begin{table*}[t]
\centering
\resizebox{0.87\textwidth}{!}{
\begin{minipage}{\textwidth}
\caption{Baseline zero-shot performance on the \textit{Full} feature set of \textit{CarbonFluxBench}. Results show the median over an ensemble of 10 models trained with different random seeds, with superscripts and subscripts denoting the 75th and 25th percentiles, respectively. Arrows indicate the direction of improvement.}
\vspace{-0.2cm}
\label{tab:full-baselines}
\centering
\begin{tabular}{ll ccc ccc ccc}
\toprule
 &  & \multicolumn{9}{c}{\textbf{Targets}} \\
\cmidrule(lr){3-11}
\textbf{Split} & \textbf{Model}
 & \multicolumn{3}{c}{\textbf{GPP}}
 & \multicolumn{3}{c}{\textbf{RECO}}
 & \multicolumn{3}{c}{\textbf{NEE}} \\
\cmidrule(lr){3-5} \cmidrule(lr){6-8} \cmidrule(lr){9-11}
 & 
 & $R^2\uparrow$ & RMSE$\downarrow$ & nMAE$\downarrow$
 & $R^2\uparrow$ & RMSE$\downarrow$ & nMAE$\downarrow$
 & $R^2\uparrow$ & RMSE$\downarrow$ & nMAE$\downarrow$ \\
\midrule
\multirow{7}{*}{IGBP}
 & LSTM        & $0.498^{0.705}_{-0.095}$ & $1.889^{2.645}_{1.138}$ & $0.477^{0.658}_{0.352}$ & $0.347^{0.715}_{-0.386}$ & $1.253^{1.966}_{0.751}$ & $0.409^{0.544}_{0.269}$ & $0.195^{0.435}_{-0.199}$ & $1.545^{2.108}_{0.816}$ & $2.123^{4.182}_{1.284}$ \\[0.5ex]
 & CT-LSTM     & $\textbf{0.561}^{0.740}_{0.111}$ & $\textbf{1.676}^{2.390}_{1.112}$ & $0.452^{0.684}_{0.323}$ & $\textbf{0.443}^{0.661}_{-0.162}$ & $1.193^{1.673}_{0.774}$ & $\textbf{0.356}^{0.526}_{0.285}$ & $0.207^{0.438}_{-0.134}$ & $1.467^{2.016}_{0.844}$ & $2.035^{3.776}_{1.340}$ \\[0.5ex]
 & GRU         & $0.420^{0.692}_{-0.009}$ & $1.794^{2.805}_{1.215}$ & $0.522^{0.648}_{0.375}$ & $0.393^{0.640}_{-0.513}$ & $1.215^{1.908}_{0.788}$ & $0.417^{0.534}_{0.294}$ & $0.150^{0.373}_{-0.077}$ & $1.444^{2.065}_{0.958}$ & $2.291^{4.202}_{1.391}$ \\[0.5ex]
 & CT-GRU      & $0.538^{0.710}_{0.217}$ & $1.802^{2.451}_{1.173}$ & $0.493^{0.673}_{0.336}$ & $0.438^{0.687}_{-0.159}$ & $1.222^{1.848}_{0.811}$ & $0.381^{0.512}_{0.289}$ & $0.159^{0.359}_{-0.368}$ & $1.588^{2.099}_{0.982}$ & $2.111^{4.345}_{1.348}$ \\[0.5ex]
 & Transformer & $0.533^{0.722}_{0.160}$ & $1.714^{2.310}_{1.169}$ & $0.457^{0.662}_{0.332}$ & $0.416^{0.674}_{-0.189}$ & $1.211^{1.744}_{0.841}$ & $0.373^{0.505}_{0.268}$ & $0.211^{0.424}_{-0.052}$ & $1.454^{2.027}_{0.868}$ & $\textbf{2.010}^{3.929}_{1.343}$ \\[0.5ex]
 & Patch-Transformer & $0.440^{0.693}_{0.072}$ & $1.822^{2.461}_{1.269}$ & $0.467^{0.683}_{0.356}$ & $0.342^{0.612}_{-0.224}$ & $1.264^{1.851}_{0.905}$ & $0.388^{0.557}_{0.311}$ & $0.107^{0.318}_{-0.214}$ & $1.564^{2.124}_{0.936}$ & $2.250^{4.139}_{1.415}$ \\[0.5ex]
 & TAM-RL      & $0.548^{0.750}_{0.087}$ & $1.683^{2.447}_{1.045}$ & $\textbf{0.443}^{0.635}_{0.332}$ & $0.416^{0.693}_{-0.207}$ & $\textbf{1.157}^{1.762}_{0.781}$ & $0.376^{0.527}_{0.272}$ & $\textbf{0.216}^{0.484}_{-0.171}$ & $\textbf{1.366}^{2.010}_{0.807}$ & $2.106^{3.948}_{1.237}$ \\[0.5ex]
\midrule
\multirow{7}{*}{K\"{o}ppen}
 & LSTM        & $0.570^{0.723}_{0.026}$ & $1.839^{2.604}_{1.248}$ & $0.476^{0.624}_{0.350}$ & $0.454^{0.697}_{-0.054}$ & $1.168^{1.779}_{0.856}$ & $0.367^{0.528}_{0.282}$ & $\textbf{0.271}^{0.425}_{-0.120}$ & $1.434^{2.159}_{1.027}$ & $2.250^{5.628}_{1.186}$ \\[0.5ex]
 & CT-LSTM     & $\textbf{0.584}^{0.754}_{0.178}$ & $\textbf{1.666}^{2.534}_{1.239}$ & $\textbf{0.413}^{0.616}_{0.312}$ & $0.515^{0.716}_{0.057}$ & $1.181^{1.783}_{0.814}$ & $0.364^{0.539}_{0.258}$ & $0.265^{0.469}_{-0.063}$ & $1.371^{1.888}_{0.979}$ & $\textbf{2.061}^{5.388}_{1.188}$ \\[0.5ex]
 & GRU         & $0.528^{0.729}_{0.158}$ & $1.780^{2.697}_{1.196}$ & $0.500^{0.620}_{0.357}$ & $0.456^{0.691}_{-0.050}$ & $1.206^{1.744}_{0.853}$ & $0.371^{0.543}_{0.273}$ & $0.228^{0.393}_{-0.219}$ & $1.422^{2.159}_{1.040}$ & $2.282^{5.384}_{1.250}$ \\[0.5ex]
 & CT-GRU      & $0.515^{0.717}_{0.169}$ & $1.808^{2.309}_{1.247}$ & $0.468^{0.657}_{0.343}$ & $0.503^{0.656}_{0.037}$ & $1.193^{1.707}_{0.820}$ & $0.376^{0.530}_{0.279}$ & $0.142^{0.392}_{-0.214}$ & $1.425^{2.074}_{1.095}$ & $2.616^{6.008}_{1.314}$ \\[0.5ex]
 & Transformer & $0.577^{0.747}_{0.218}$ & $1.828^{2.501}_{1.163}$ & $0.445^{0.641}_{0.335}$ & $0.483^{0.686}_{0.018}$ & $1.196^{1.741}_{0.902}$ & $0.369^{0.516}_{0.264}$ & $0.266^{0.479}_{-0.031}$ & $\textbf{1.322}^{1.825}_{0.968}$ & $2.272^{5.518}_{1.155}$ \\[0.5ex]
 & Patch-Transformer & $0.458^{0.696}_{0.067}$ & $1.949^{2.637}_{1.372}$ & $0.495^{0.659}_{0.367}$ & $0.485^{0.660}_{-0.019}$ & $1.293^{1.734}_{0.940}$ & $0.375^{0.550}_{0.311}$ & $0.123^{0.305}_{-0.199}$ & $1.570^{2.117}_{1.028}$ & $2.453^{6.208}_{1.328}$ \\[0.5ex]
 & TAM-RL      & $0.566^{0.748}_{0.096}$ & $1.740^{2.376}_{1.252}$ & $0.450^{0.635}_{0.340}$ & $\textbf{0.546}^{0.699}_{-0.060}$ & $\textbf{1.162}^{1.695}_{0.878}$ & $\textbf{0.353}^{0.535}_{0.279}$ & $0.243^{0.443}_{-0.071}$ & $1.380^{1.860}_{1.005}$ & $2.276^{5.314}_{1.168}$ \\[0.5ex]
\bottomrule
      \vspace{-0.7cm}
\end{tabular}
 
\end{minipage}
}
\end{table*}

\RaggedRight
\section{Extended Related Work} \label{app:related}

This appendix provides a comprehensive review of (1) carbon flux upscaling methods and operational products, and (2) transfer learning benchmarks in machine learning, expanding on the brief overview in Section~3.

\subsection{Carbon Flux Upscaling: Methods and Products}
The primary goal of carbon flux upscaling is to produce improved gridded estimates of terrestrial carbon exchange for climate modeling and decision-making. Given the extensive existing literature, we do not attempt to exhaustively review all prior efforts, instead, we note that most approaches follow a common strategy. Specifically, they rely on target variables derived from global eddy covariance (EC) networks such as FLUXNET \cite{baldocchi2001fluxnet} and AmeriFlux \cite{novick2018ameriflux}, and estimate gross primary production (GPP; downward carbon flux), ecosystem respiration (RECO; upward carbon flux), and net ecosystem exchange (NEE; net balance), either individually or jointly in a multitask learning setting. These quantities are typically modeled as functions of remotely sensed vegetation indicators and gridded meteorological drivers at the target locations.

One of the most significant and earliest efforts in this domain is FLUXCOM \cite{jung2020scaling} and its recent successor FLUXCOM-X-BASE \cite{nelson2024x}. The FLUXCOM framework integrates multiple algorithmic families, including kernel methods, neural networks, tree-based models, and regression splines, and is trained under two configurations: (a) a Remote Sensing (RS) setup using only multispectral remote sensing observations at 8-day temporal and 0.0833° spatial resolution, and (b) an RS+Meteo setup using additional daily meteorological variables at 0.5° spatial resolution. The newer FLUXCOM-X-BASE leverages the Extreme Gradient Boosting (XGBoost) \cite{chen2016xgboost} architecture combined with MODIS and climate forcings to generate global flux estimates at hourly temporal and 0.05° spatial resolution, achieving notable improvements over the original FLUXCOM product.

Another recent dataset, MetaFlux \cite{nathaniel2023metaflux}, introduces a meta-learning framework for carbon flux upscaling using deep learning architectures. The model operates in two stages: meta-training on base tasks with abundant samples (e.g., the sites in continental climates), followed by meta-updating on data-sparse tasks (e.g., tropical sites). Like FLUXCOM, MetaFlux relies on meteorological and MODIS-derived features, providing daily and monthly flux estimates at 0.25° spatial and 1d temporal resolution for 2001–2021.

Later, \cite{yuan2025global} proposed an alternative transfer learning strategy based on the XGBoost architecture. Their approach first trains a universal base model using all available samples and predictors, followed by fine-tuning specialized models on subsets of data corresponding to distinct plant functional types (PFTs). This framework was used to produce GloFlux, a global carbon flux dataset at 0.1° spatial and monthly temporal resolution.

In parallel, knowledge-guided machine learning (KGML) approaches have gained increasing attention in recent years. A study by \cite{fan2025estimating} applied a transformer-based model \cite{vaswani2017attention} to upscale carbon fluxes over the United States at 500 m spatial and annual temporal resolution. The proposed method incorporated a physics-inspired loss function enforcing the carbon balance constraint, $NEE = RECO - GPP$. The authors reported that introducing this constraint led to RMSE reductions of 1.5–3.5\%, demonstrating the potential benefits of embedding domain knowledge into learning objectives.

Despite these methodological advances, existing studies differ substantially in their selection of EC sites, target variables, feature sets, evaluation protocols, and performance metrics. This heterogeneity makes it difficult to assess whether reported improvements reflect genuine methodological progress or are artifacts of dataset and experimental design choices. Therefore, a standardized benchmark that bridges the machine learning and carbon flux research communities is essential for enabling transparent comparison and accelerating future progress.

\subsection{Transfer Learning Benchmarks in Machine Learning}

While transfer learning has become a cornerstone of modern machine learning, existing benchmarks exhibit notable gaps that limit progress in certain application domains.

\textbf{Classification-dominated landscape:} The majority of transfer learning benchmarks focus on classification tasks. In computer vision, unprecedented growth has been driven by standardized benchmarks \cite{lecun2010mnist, krizhevsky2009learning, deng2009imagenet} that enable systematic methodological comparison. Domain adaptation benchmarks such as Office-31 \cite{saenko2010adapting}, Office-Home \cite{venkateswara2017deep}, VisDA \cite{peng2017visda}, and DomainNet \cite{peng2019moment} evaluate cross-domain image classification. In NLP, benchmarks like GLUE \cite{wang2018glue}, and SQuAD \cite{rajpurkar2016squad} primarily assess classification and span prediction tasks.

\textbf{Limited regression benchmarks:} Transfer learning for regression remains underexplored in the benchmarking literature. Regression under domain shift presents distinct challenges absent in classification: continuous predictions across varying scales, preservation of calibration, and sensitivity to distributional changes in target variables. The few existing regression benchmarks (e.g., UCI datasets \cite{dua2017uci} with artificial domain shifts) lack the scale, diversity, and real-world complexity needed to drive methodological innovation. This is especially true and more difficult and complex when considering time series regression.

\textbf{Time series adds another layer of complexity:} While some benchmarks address time series forecasting, including the Monash Time Series Forecasting Archive \cite{godahewa2021monash}, Informer \cite{zhou2021informer}, and Autoformer \cite{wu2021autoformer}, they typically evaluate temporal generalization (forecasting future values at known locations) rather than spatial transfer. Time series regression with spatial transfer is substantially more challenging because models must simultaneously (1) capture temporal dependencies and seasonal patterns, (2) learn representations that transfer across spatial domains with different temporal dynamics, and (3) handle regime shifts where the same temporal patterns may have different meanings in different locations (e.g., temperature fluctuations in tropical vs. polar ecosystems).

\textbf{Absence of spatial transfer evaluation:} No existing benchmark to our knowledge systematically evaluates spatial generalization i.e., the ability to transfer models to geographically distinct locations with different environmental conditions, ecosystem properties, or climatic regimes. This is a fundamental challenge in Earth system sciences \cite{reichstein2019deep, camps2021deep}, ecology \cite{karpatne2017machine}, agriculture \cite{kamilaris2018deep}, and public health \cite{shang2021regional}, where observations are spatially sparse and expensive to collect, yet predictions are needed globally.

\textbf{Need for scientific domain benchmarks:} Scientific applications offer natural testbeds for spatial transfer learning: they feature genuine spatial heterogeneity (not artificial), provide interpretable feature spaces grounded in physical processes, and have clear societal impact. Yet they remain disconnected from the ML benchmarking ecosystem \cite{willard2022integrating, karpatne2017machine}. Bridging this gap could accelerate both methodological innovation in transfer learning and scientific discovery in domain applications.

CarbonFluxBench addresses these gaps by providing the first benchmark for spatial transfer learning in time series regression, grounded in a scientifically important application with real-world impact.

\section{Eddy Covariance Sites}

\begin{figure}[h]
    \centering
    \includegraphics[width=1\linewidth]{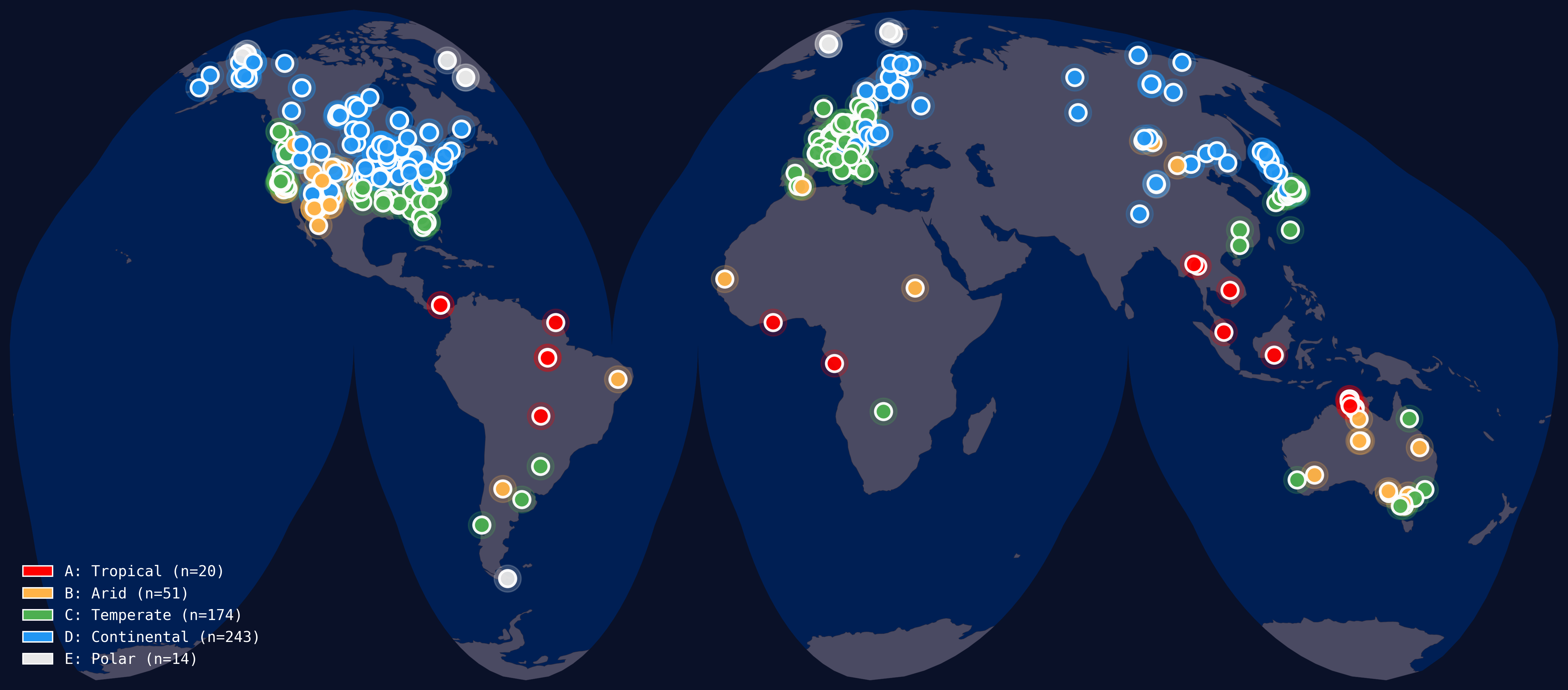}
    \caption{Geographic distribution of 474 eddy covariance sites in CarbonFluxBench, colored by Köppen climate classification.}
    \label{fig:sites}
\end{figure}

Figure~\ref{fig:sites} shows the geographic distribution of all 567 sites included in CarbonFluxBench, colored by Köppen climate classification. The spatial heterogeneity and concentration in North America and Europe motivates our dual stratification strategy. 

\section{Training Protocol Details}
\label{app:training}

\subsection{Loss Function}

To guide neural network training, we employ a composite loss function that has been shown to improve carbon flux prediction in prior work \cite{rozanov2025tamrl_carbon}. This loss incorporates data quality weighting, class balancing, and physical consistency constraints:
\[
\mathcal{L} = \text{MSE} \cdot w_{\text{qc}} \cdot w_{\text{igbp}} \cdot w_{\text{koppen}} + \alpha \cdot \mathcal{L}_{\text{flux}}, \quad \alpha = 0.1
\]

The loss components are defined as follows:
\begin{itemize}
    \item \textbf{Quality weighting ($w_{\text{qc}}$):} Eddy covariance targets vary in reliability due to gap-filling and filtering procedures. We weight samples by their continuous quality flag (ranging from 0 to 1), enabling partial utilization of lower-quality observations instead of binary exclusion.
    
    \item \textbf{Class balance weights ($w_{\text{igbp}}$, $w_{\text{koppen}}$):} Flux tower coverage is highly uneven across land cover and climate classes (Table~\ref{tab:splits}). We apply inverse-frequency weighting to increase loss contributions from underrepresented ecosystem and climate types.
    
    \item \textbf{Flux balance constraint ($\mathcal{L}_{\text{flux}}$):} To enforce physical consistency between targets, we penalize violations of the carbon balance relationship $\text{NEE} = \text{RECO} - \text{GPP}$.
\end{itemize}

\section{Sites} \label{app:sites}
\begin{table*}[htbp]
\centering
\caption{AmeriFlux sites used in this study and their dataset citations. We cite individual AmeriFlux sites as it is required by the data policy.}
\footnotesize
\setlength{\tabcolsep}{3pt}
\begin{tabular}{lllllllll}
\toprule
\multicolumn{9}{c}{\textbf{AmeriFlux}} \\
\midrule
\texttt{AR-CCg}~\cite{AR-CCg_AMF} & \texttt{AR-TF1}~\cite{AR-TF1_AMF} & \texttt{BR-CST}~\cite{BR-CST_AMF} & \texttt{BR-Npw}~\cite{BR-Npw_AMF} & \texttt{CA-ARB}~\cite{CA-ARB_AMF} & \texttt{CA-ARF}~\cite{CA-ARF_AMF} & \texttt{CA-BOU}~\cite{CA-BOU_AMF} & \texttt{CA-Ca1}~\cite{CA-Ca1_AMF} & \texttt{CA-Ca2}~\cite{CA-Ca2_AMF} \\
\texttt{CA-Cbo}~\cite{CA-Cbo_AMF} & \texttt{CA-CF1}~\cite{CA-CF1_AMF} & \texttt{CA-DB2}~\cite{CA-DB2_AMF} & \texttt{CA-DBB}~\cite{CA-DBB_AMF} & \texttt{CA-EM1}~\cite{CA-EM1_AMF} & \texttt{CA-ER1}~\cite{CA-ER1_AMF} & \texttt{CA-HPC}~\cite{CA-HPC_AMF} & \texttt{CA-LP1}~\cite{CA-LP1_AMF} & \texttt{CA-MA1}~\cite{CA-MA1_AMF} \\
\texttt{CA-MA2}~\cite{CA-MA2_AMF} & \texttt{CA-MA3}~\cite{CA-MA3_AMF} & \texttt{CA-Qc2}~\cite{CA-Qc2_AMF} & \texttt{CA-SCB}~\cite{CA-SCB_AMF} & \texttt{CA-SCC}~\cite{CA-SCC_AMF} & \texttt{CA-TP1}~\cite{CA-TP1_AMF} & \texttt{CA-TP3}~\cite{CA-TP3_AMF} & \texttt{CA-TPD}~\cite{CA-TPD_AMF} & \texttt{CL-SDF}~\cite{CL-SDF_AMF} \\
\texttt{MX-Tes}~\cite{MX-Tes_AMF} & \texttt{US-A32}~\cite{US-A32_AMF} & \texttt{US-A74}~\cite{US-A74_AMF} & \texttt{US-Akn}~\cite{US-Akn_AMF} & \texttt{US-ALQ}~\cite{US-ALQ_AMF} & \texttt{US-ARM}~\cite{US-ARM_AMF} & \texttt{US-ASH}~\cite{US-ASH_AMF} & \texttt{US-ASM}~\cite{US-ASM_AMF} & \texttt{US-Bar}~\cite{US-Bar_AMF} \\
\texttt{US-Bi1}~\cite{US-Bi1_AMF} & \texttt{US-Bi2}~\cite{US-Bi2_AMF} & \texttt{US-BRG}~\cite{US-BRG_AMF} & \texttt{US-BZB}~\cite{US-BZB_AMF} & \texttt{US-BZF}~\cite{US-BZF_AMF} & \texttt{US-BZo}~\cite{US-BZo_AMF} & \texttt{US-BZS}~\cite{US-BZS_AMF} & \texttt{US-CdM}~\cite{US-CdM_AMF} & \texttt{US-CF1}~\cite{US-CF1_AMF} \\
\texttt{US-CF2}~\cite{US-CF2_AMF} & \texttt{US-CF3}~\cite{US-CF3_AMF} & \texttt{US-CF4}~\cite{US-CF4_AMF} & \texttt{US-CGG}~\cite{US-CGG_AMF} & \texttt{US-CS1}~\cite{US-CS1_AMF} & \texttt{US-CS2}~\cite{US-CS2_AMF} & \texttt{US-CS3}~\cite{US-CS3_AMF} & \texttt{US-CS4}~\cite{US-CS4_AMF} & \texttt{US-CS5}~\cite{US-CS5_AMF} \\
\texttt{US-CS6}~\cite{US-CS6_AMF} & \texttt{US-CS8}~\cite{US-CS8_AMF} & \texttt{US-Cst}~\cite{US-Cst_AMF} & \texttt{US-DFC}~\cite{US-DFC_AMF} & \texttt{US-Dmg}~\cite{US-Dmg_AMF} & \texttt{US-DS3}~\cite{US-DS3_AMF} & \texttt{US-EDN}~\cite{US-EDN_AMF} & \texttt{US-EML}~\cite{US-EML_AMF} & \texttt{US-Fcr}~\cite{US-Fcr_AMF} \\
\texttt{US-Fmf}~\cite{US-Fmf_AMF} & \texttt{US-Fuf}~\cite{US-Fuf_AMF} & \texttt{US-Fwf}~\cite{US-Fwf_AMF} & \texttt{US-GLE}~\cite{US-GLE_AMF} & \texttt{US-Ha1}~\cite{US-Ha1_AMF} & \texttt{US-HB1}~\cite{US-HB1_AMF} & \texttt{US-HB2}~\cite{US-HB2_AMF} & \texttt{US-HB3}~\cite{US-HB3_AMF} & \texttt{US-Hn2}~\cite{US-Hn2_AMF} \\
\texttt{US-Hn3}~\cite{US-Hn3_AMF} & \texttt{US-Ho1}~\cite{US-Ho1_AMF} & \texttt{US-Ho2}~\cite{US-Ho2_AMF} & \texttt{US-HWB}~\cite{US-HWB_AMF} & \texttt{US-ICh}~\cite{US-ICh_AMF} & \texttt{US-ICs}~\cite{US-ICs_AMF} & \texttt{US-ICt}~\cite{US-ICt_AMF} & \texttt{US-Jo1}~\cite{US-Jo1_AMF} & \texttt{US-Jo2}~\cite{US-Jo2_AMF} \\
\texttt{US-KFS}~\cite{US-KFS_AMF} & \texttt{US-KLS}~\cite{US-KLS_AMF} & \texttt{US-Kon}~\cite{US-Kon_AMF} & \texttt{US-KS1}~\cite{US-KS1_AMF} & \texttt{US-KS3}~\cite{US-KS3_AMF} & \texttt{US-LS2}~\cite{US-LS2_AMF} & \texttt{US-Me2}~\cite{US-Me2_AMF} & \texttt{US-Me6}~\cite{US-Me6_AMF} & \texttt{US-MMS}~\cite{US-MMS_AMF} \\
\texttt{US-Mo1}~\cite{US-Mo1_AMF} & \texttt{US-Mo2}~\cite{US-Mo2_AMF} & \texttt{US-Mo3}~\cite{US-Mo3_AMF} & \texttt{US-MOz}~\cite{US-MOz_AMF} & \texttt{US-Mpj}~\cite{US-Mpj_AMF} & \texttt{US-Myb}~\cite{US-Myb_AMF} & \texttt{US-NC1}~\cite{US-NC1_AMF} & \texttt{US-NC3}~\cite{US-NC3_AMF} & \texttt{US-NC4}~\cite{US-NC4_AMF} \\
\texttt{US-Ne1}~\cite{US-Ne1_AMF} & \texttt{US-NGB}~\cite{US-NGB_AMF} & \texttt{US-NGC}~\cite{US-NGC_AMF} & \texttt{US-NR1}~\cite{US-NR1_AMF} & \texttt{US-ONA}~\cite{US-ONA_AMF} & \texttt{US-ORv}~\cite{US-ORv_AMF} & \texttt{US-OWC}~\cite{US-OWC_AMF} & \texttt{US-PFb}~\cite{US-PFb_AMF} & \texttt{US-PFc}~\cite{US-PFc_AMF} \\
\texttt{US-PFd}~\cite{US-PFd_AMF} & \texttt{US-PFk}~\cite{US-PFk_AMF} & \texttt{US-PFL}~\cite{US-PFL_AMF} & \texttt{US-PFm}~\cite{US-PFm_AMF} & \texttt{US-PFn}~\cite{US-PFn_AMF} & \texttt{US-PFp}~\cite{US-PFp_AMF} & \texttt{US-PFq}~\cite{US-PFq_AMF} & \texttt{US-PFt}~\cite{US-PFt_AMF} & \texttt{US-Pnp}~\cite{US-Pnp_AMF} \\
\texttt{US-PSH}~\cite{US-PSH_AMF} & \texttt{US-PSL}~\cite{US-PSL_AMF} & \texttt{US-RGA}~\cite{US-RGA_AMF} & \texttt{US-RGB}~\cite{US-RGB_AMF} & \texttt{US-RGF}~\cite{US-RGF_AMF} & \texttt{US-RGo}~\cite{US-RGo_AMF} & \texttt{US-RGW}~\cite{US-RGW_AMF} & \texttt{US-Rls}~\cite{US-Rls_AMF} & \texttt{US-Rms}~\cite{US-Rms_AMF} \\
\texttt{US-Ro1}~\cite{US-Ro1_AMF} & \texttt{US-Ro2}~\cite{US-Ro2_AMF} & \texttt{US-Ro3}~\cite{US-Ro3_AMF} & \texttt{US-Ro4}~\cite{US-Ro4_AMF} & \texttt{US-Ro5}~\cite{US-Ro5_AMF} & \texttt{US-Ro6}~\cite{US-Ro6_AMF} & \texttt{US-Rpf}~\cite{US-Rpf_AMF} & \texttt{US-RRC}~\cite{US-RRC_AMF} & \texttt{US-Rwe}~\cite{US-Rwe_AMF} \\
\texttt{US-Rwf}~\cite{US-Rwf_AMF} & \texttt{US-Rws}~\cite{US-Rws_AMF} & \texttt{US-Seg}~\cite{US-Seg_AMF} & \texttt{US-Ses}~\cite{US-Ses_AMF} & \texttt{US-Sne}~\cite{US-Sne_AMF} & \texttt{US-Snf}~\cite{US-Snf_AMF} & \texttt{US-SP1}~\cite{US-SP1_AMF} & \texttt{US-SRG}~\cite{US-SRG_AMF} & \texttt{US-SRM}~\cite{US-SRM_AMF} \\
\texttt{US-Srr}~\cite{US-Srr_AMF} & \texttt{US-SRS}~\cite{US-SRS_AMF} & \texttt{US-SSH}~\cite{US-SSH_AMF} & \texttt{US-StJ}~\cite{US-StJ_AMF} & \texttt{US-Syv}~\cite{US-Syv_AMF} & \texttt{US-Ton}~\cite{US-Ton_AMF} & \texttt{US-Tw1}~\cite{US-Tw1_AMF} & \texttt{US-Tw3}~\cite{US-Tw3_AMF} & \texttt{US-Tw4}~\cite{US-Tw4_AMF} \\
\texttt{US-Tw5}~\cite{US-Tw5_AMF} & \texttt{US-UC1}~\cite{US-UC1_AMF} & \texttt{US-UC2}~\cite{US-UC2_AMF} & \texttt{US-UM3}~\cite{US-UM3_AMF} & \texttt{US-UMB}~\cite{US-UMB_AMF} & \texttt{US-UMd}~\cite{US-UMd_AMF} & \texttt{US-Var}~\cite{US-Var_AMF} & \texttt{US-Vcm}~\cite{US-Vcm_AMF} & \texttt{US-Vcp}~\cite{US-Vcp_AMF} \\
\texttt{US-Whs}~\cite{US-Whs_AMF} & \texttt{US-Wjs}~\cite{US-Wjs_AMF} & \texttt{US-Wkg}~\cite{US-Wkg_AMF} & \texttt{US-xAB}~\cite{US-xAB_AMF} & \texttt{US-xAE}~\cite{US-xAE_AMF} & \texttt{US-xBA}~\cite{US-xBA_AMF} & \texttt{US-xBL}~\cite{US-xBL_AMF} & \texttt{US-xBN}~\cite{US-xBN_AMF} & \texttt{US-xBR}~\cite{US-xBR_AMF} \\
\texttt{US-xCL}~\cite{US-xCL_AMF} & \texttt{US-xCP}~\cite{US-xCP_AMF} & \texttt{US-xDC}~\cite{US-xDC_AMF} & \texttt{US-xDJ}~\cite{US-xDJ_AMF} & \texttt{US-xDL}~\cite{US-xDL_AMF} & \texttt{US-xDS}~\cite{US-xDS_AMF} & \texttt{US-xGR}~\cite{US-xGR_AMF} & \texttt{US-xHA}~\cite{US-xHA_AMF} & \texttt{US-xHE}~\cite{US-xHE_AMF} \\
\texttt{US-xJE}~\cite{US-xJE_AMF} & \texttt{US-xJR}~\cite{US-xJR_AMF} & \texttt{US-xKA}~\cite{US-xKA_AMF} & \texttt{US-xKZ}~\cite{US-xKZ_AMF} & \texttt{US-xMB}~\cite{US-xMB_AMF} & \texttt{US-xML}~\cite{US-xML_AMF} & \texttt{US-xNG}~\cite{US-xNG_AMF} & \texttt{US-xNQ}~\cite{US-xNQ_AMF} & \texttt{US-xRM}~\cite{US-xRM_AMF} \\
\texttt{US-xSB}~\cite{US-xSB_AMF} & \texttt{US-xSC}~\cite{US-xSC_AMF} & \texttt{US-xSE}~\cite{US-xSE_AMF} & \texttt{US-xSJ}~\cite{US-xSJ_AMF} & \texttt{US-xSL}~\cite{US-xSL_AMF} & \texttt{US-xSR}~\cite{US-xSR_AMF} & \texttt{US-xST}~\cite{US-xST_AMF} & \texttt{US-xTA}~\cite{US-xTA_AMF} & \texttt{US-xTL}~\cite{US-xTL_AMF} \\
\texttt{US-xTR}~\cite{US-xTR_AMF} & \texttt{US-xUK}~\cite{US-xUK_AMF} & \texttt{US-xUN}~\cite{US-xUN_AMF} & \texttt{US-xWD}~\cite{US-xWD_AMF} & \texttt{US-xYE}~\cite{US-xYE_AMF} & \texttt{US-YK2}~\cite{US-YK2_AMF} &  &  &  \\
\bottomrule
\end{tabular}
\label{tab:flux_sites_amf}
\end{table*}
